\documentclass{article}

\usepackage{microtype}
\usepackage{graphicx}
\usepackage{subcaption}
\usepackage{booktabs} 
\usepackage{enumitem}
\usepackage{graphicx}
\usepackage{wrapfig}
\usepackage{booktabs}
\usepackage{multirow}
\usepackage[table]{xcolor}

\usepackage{hyperref}

\newcommand*{\eg}{\emph{e.g.},\@\xspace}
\newcommand*{\ie}{\emph{i.e.},\@\xspace}

\usepackage[accepted]{icml2026}

\usepackage{amsmath}
\usepackage{amssymb}
\usepackage{mathtools}
\usepackage{amsthm}
\usepackage{xspace}

\usepackage{listings}
\usepackage{xcolor}

\usepackage[most]{tcolorbox}
\usepackage{fvextra}    %
\usepackage{microtype}

\definecolor{IllinoisOrange}{HTML}{FF5F05}
\definecolor{IllinoisBlue}{HTML}{13294B}
\newcommand{\logoicon}{%
  \raisebox{-0.20\height}{\includegraphics[height=1.2em]{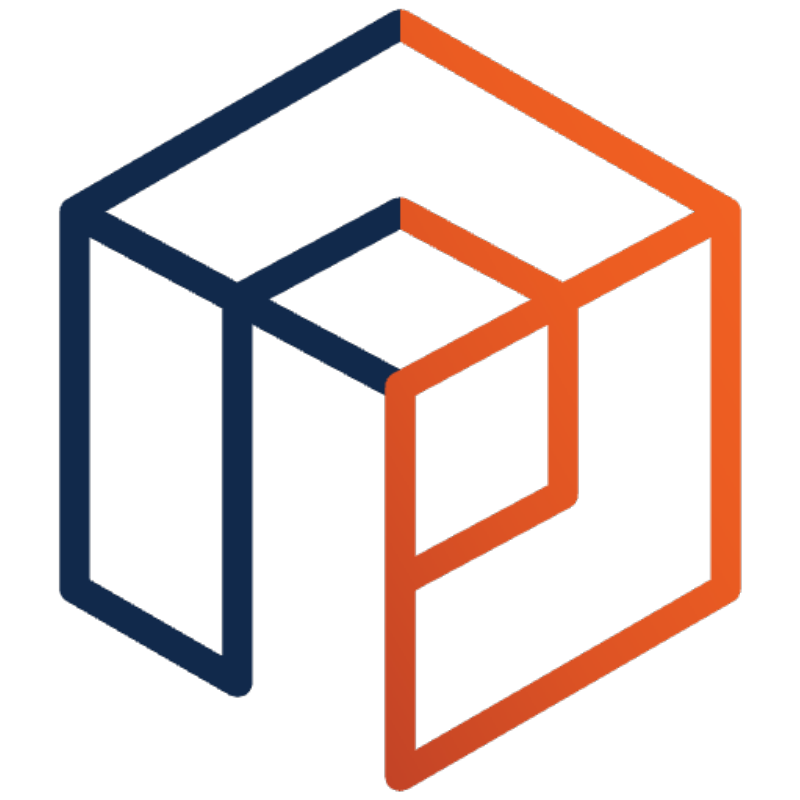}}%
}

\lstdefinelanguage{prompt}{
  morestring=[b]",
  morecomment=[l]{//},
}
\usepackage[capitalize,noabbrev]{cleveref}

\theoremstyle{plain}

\theoremstyle{definition}

\theoremstyle{remark}

\usepackage{enumitem}
\setlist[itemize]{leftmargin=*}

\usepackage[textsize=tiny]{todonotes}

\newcommand{\il}[1]{\textcolor{orange}{[IL: #1]}}

\newcommand{\modelnamenc}{UniDFlow\xspace}
\newcommand{\dponamenc}{mRef-DPO\xspace}

\icmltitlerunning{Best of Both Worlds: Multimodal Reasoning and Generation via Unified Discrete Flow Matching}

\begin{document}

\twocolumn[
\icmltitle{Best of Both Worlds: Multimodal Reasoning and Generation \\via Unified Discrete Flow Matching}



  \icmlsetsymbol{equal}{*}

  \begin{icmlauthorlist}
    \icmlauthor{Onkar Susladkar}{yyy}
    \icmlauthor{Tushar Prakash}{abc}
    \icmlauthor{Gayatri Deshmukh}{yyy}
    \icmlauthor{Kiet A. Nguyen}{yyy}
    \icmlauthor{Jiaxun Zhang}{yyy}
    \icmlauthor{Adheesh Juvekar}{yyy}
    \icmlauthor{Tianshu Bao}{comp}
    \icmlauthor{Lin Chai}{comp}
    \icmlauthor{Sparsh Mittal}{def}
    \icmlauthor{Inderjit Dhillon}{comp,scz}
    \icmlauthor{Ismini Lourentzou}{yyy}
  \end{icmlauthorlist}

  \icmlaffiliation{yyy}{University of Illinois Urbana-Champaign}
  \icmlaffiliation{comp}{Google Research}
  \icmlaffiliation{abc}{Sony Research, India}
  \icmlaffiliation{def}{Indian Institute of Technology, Roorkee}
  \icmlaffiliation{scz}{University of Texas at Austin}
  
  \icmlcorrespondingauthor{Onkar Susladkar}{onkarks2@illinois.edu, lourent2@illinois.edu}

  \icmlkeywords{Machine Learning, ICML}

\vskip 0.15in 

  {
      \centering
      \includegraphics[width=\textwidth]{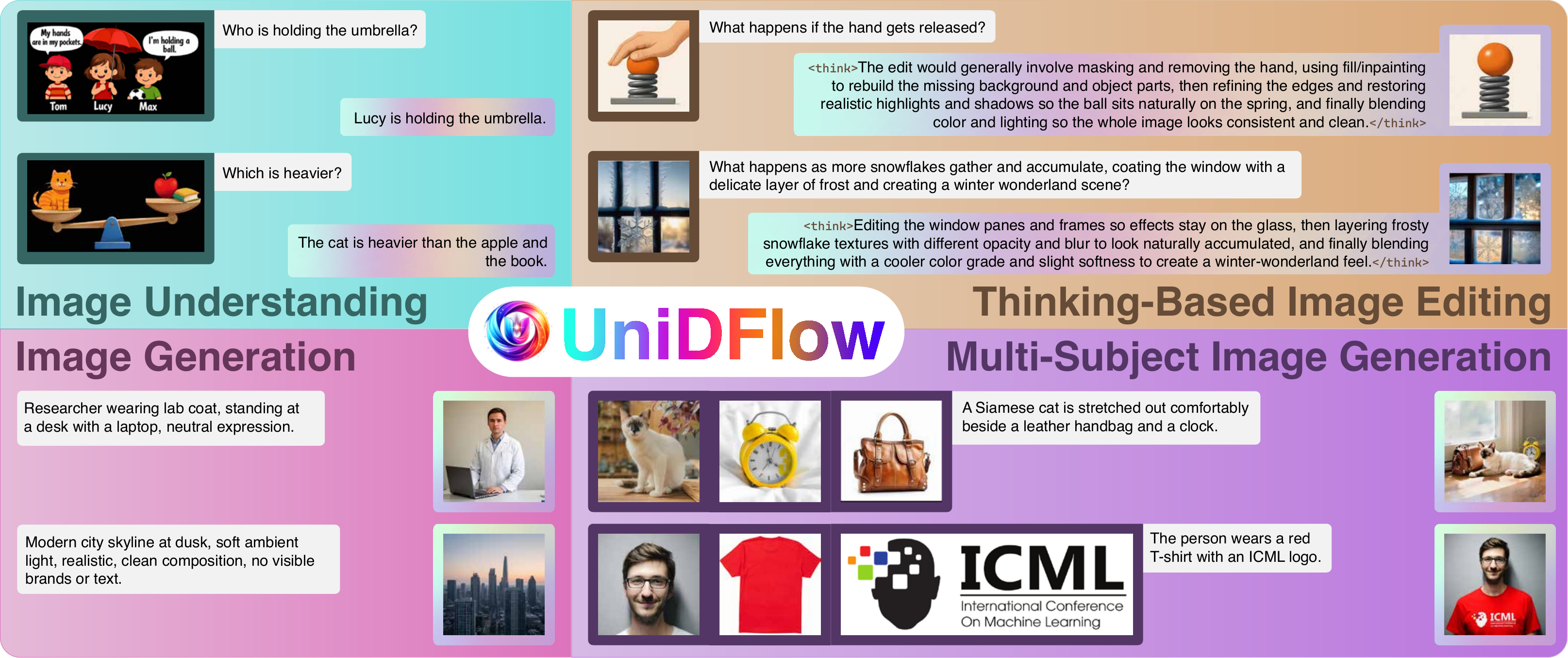}
     \captionof{figure}{We propose \modelnamenc, an unified multimodal diffusion framework that supports image understanding, generation, and thinking-based editing. The model performs visual reasoning for question answering, produces high-quality text-to-image generations across diverse scenes and subjects, and enables instruction-driven, multi-step image editing through structured reasoning.}
      \label{fig:teaser}
  }

  \vskip 0.2in
]



\printAffiliationsAndNotice{}  
\newcommand{\FigIntro}{
\begin{figure}[t!]
\centering
\includegraphics[width=0.9\linewidth]{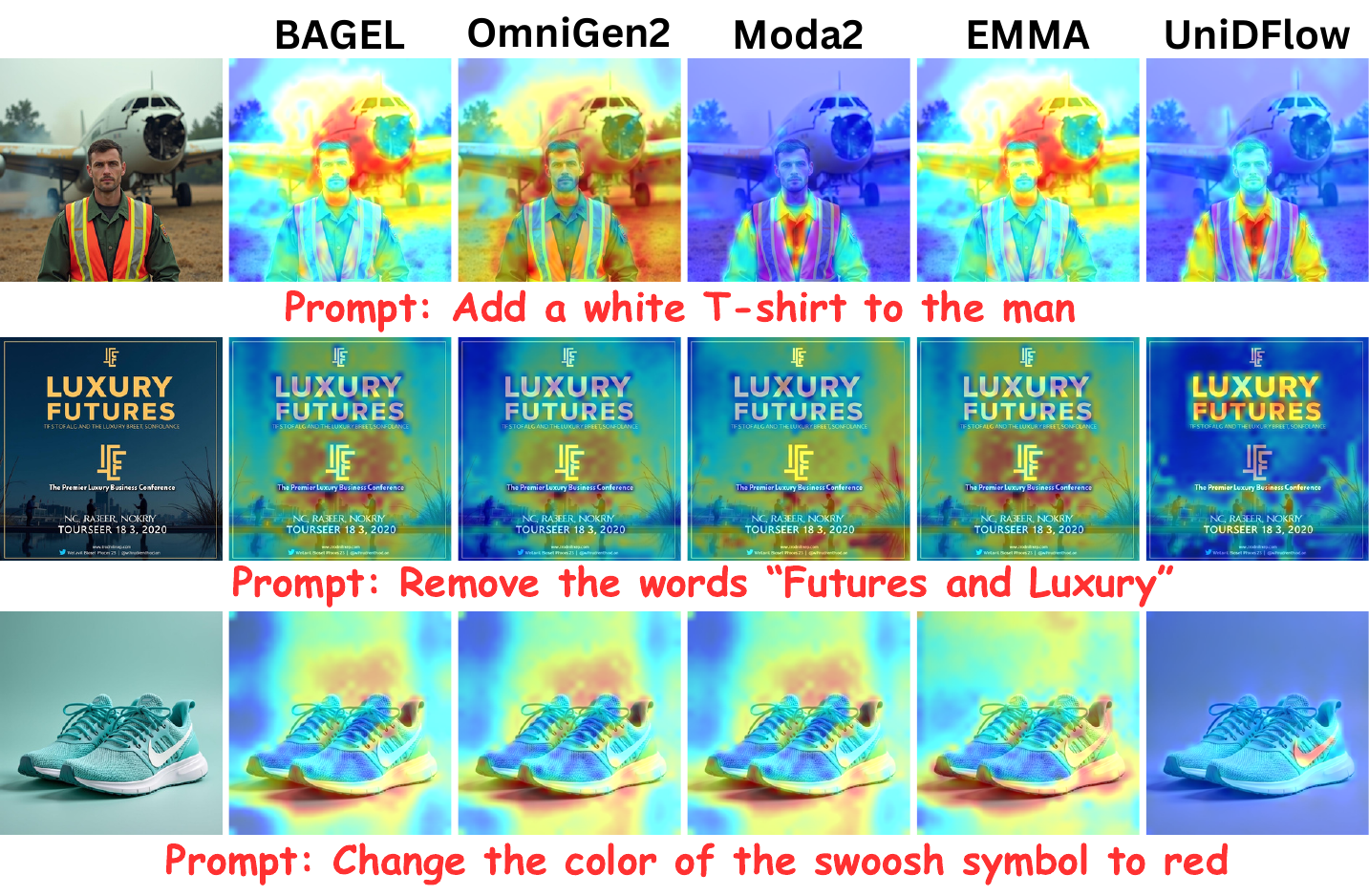}
\vspace{-0.15cm}
    \caption{Instruction-guided editing attention maps showing \modelnamenc more precisely focuses on relevant regions than prior models.\looseness-1}
    \label{fig:activation_map}
\vspace{-0.1cm}
\end{figure}
}

\newcommand{\FigArch}{
\begin{figure}[t!]
    \centering    \includegraphics[width=0.93\linewidth]{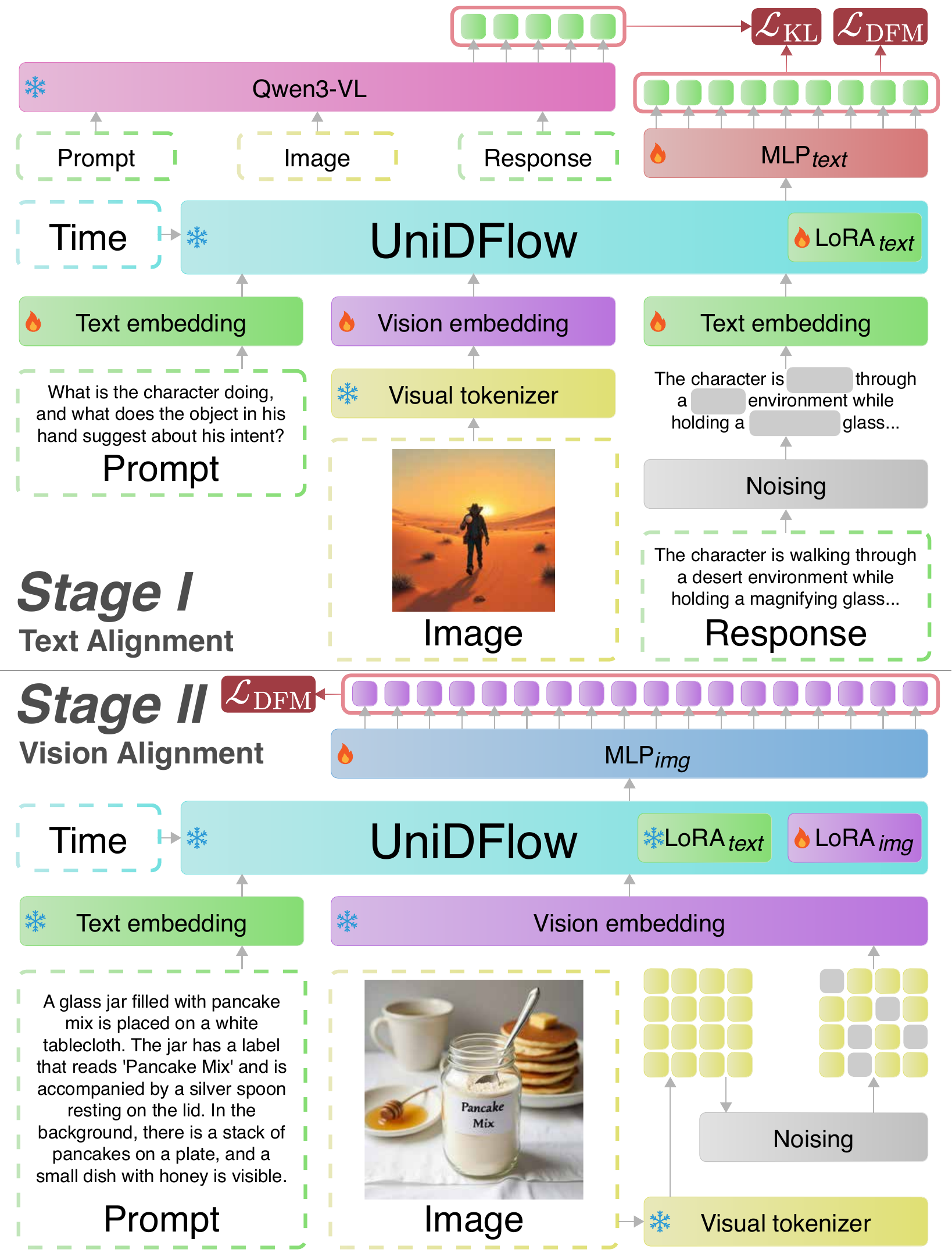}
    \caption{Overview of Stage I (understanding via text alignment) and Stage II (generation via vision alignment) of \modelnamenc{}.}
    \label{fig:overview_stage12}
\end{figure}}

\newcommand{\FigArchStageThree}{
\begin{figure*}[t!]
    \centering    \includegraphics[width=0.99\linewidth]{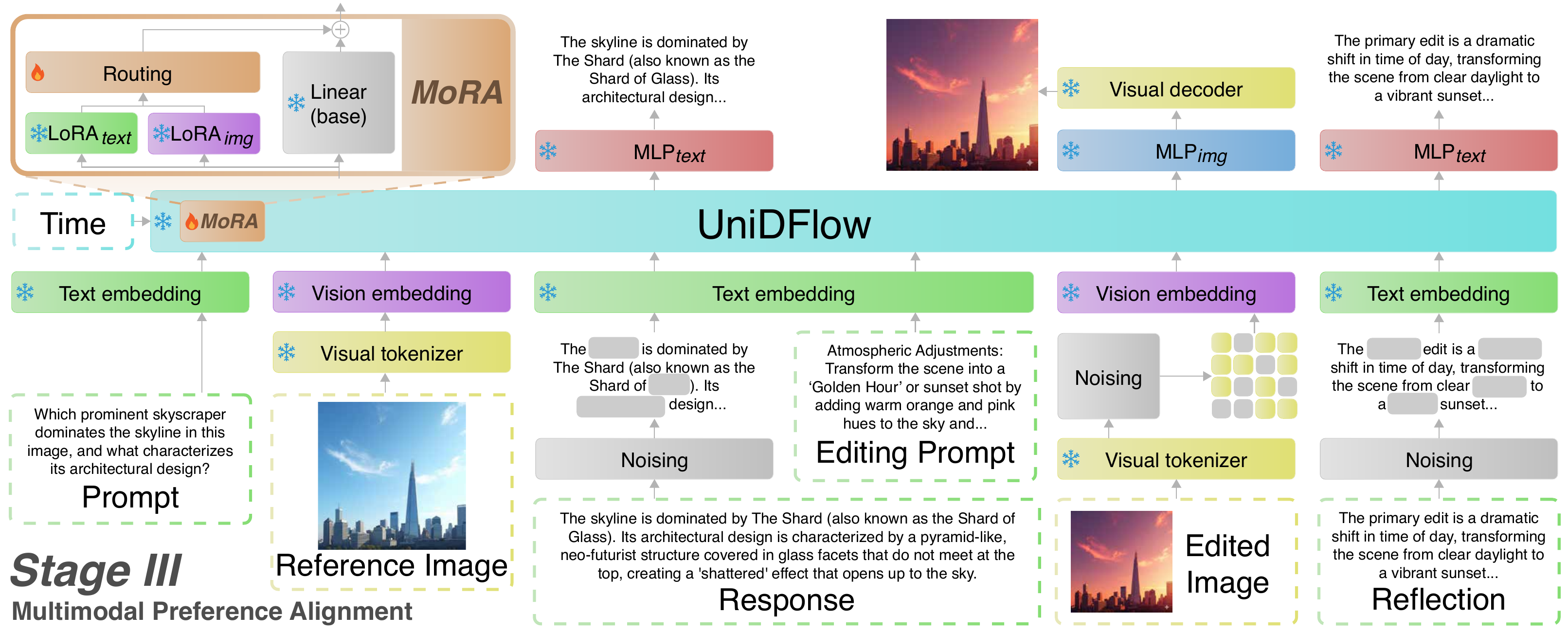}
    \caption{Stage III of \modelnamenc{}: reference-based multimodal preference alignment for improved faithfulness, controllability, and editing.}
    \label{fig:overview_stage3}
    \vspace{-0.2cm}
\end{figure*}}

\newcommand{\Figeditingintelli}
{\begin{figure}[htbp]
\centering\includegraphics[width=.95\linewidth]{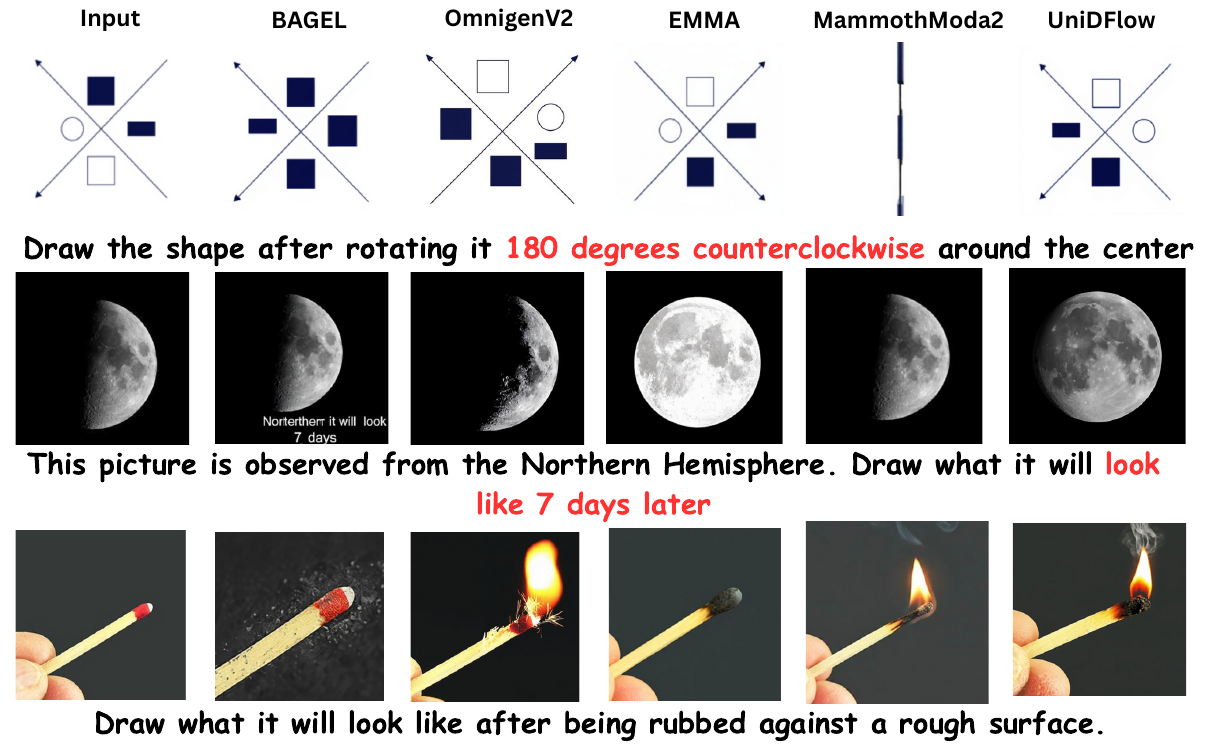}
    \caption{Reasoning-driven image editing, highlighting
temporal, geometric, and physical transformations handled by
\modelnamenc.}
    \label{fig:edit_intelligent}
\vspace{0.1cm}
\end{figure}}

\newcommand{\Figgenerationsimple}
{\begin{figure*}[t!]
\centering\includegraphics[width=.92\linewidth]{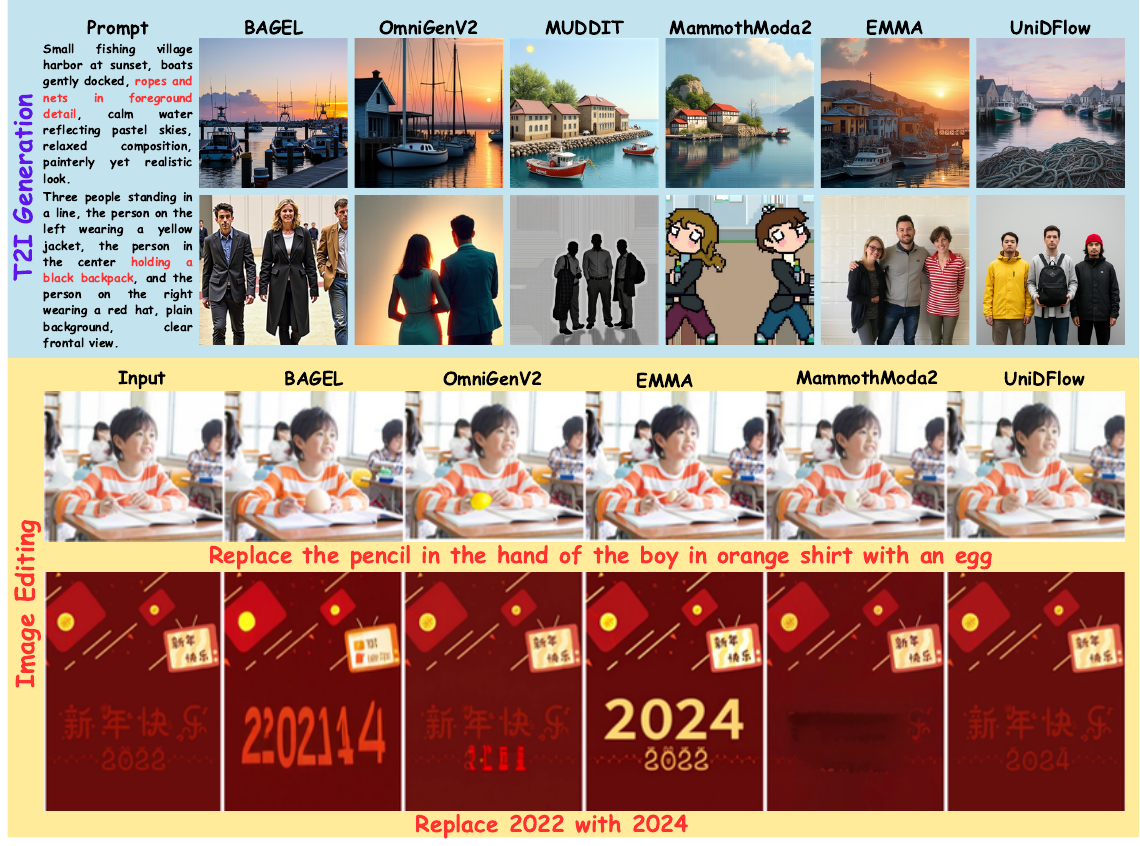}
\vspace{-0.2cm}
    \caption{Qualitative comparison of compositional text-to-image generation and editing.
Prompts require precise grounding of attributes and spatial
relations (red text). \modelnamenc consistently adheres to these
constraints while maintaining realistic structure and visual fidelity,
outperforming prior unified baselines in fine-grained prompt alignment. More results can be found in Appendix~\ref{sec:apn_mr}.}
    \label{fig:generation_1}
    \vspace{-0.2cm}
\end{figure*}}

\newcommand{\Figgenemultiobject}
{\begin{figure}[t!]
\centering\includegraphics[width=.95\columnwidth]{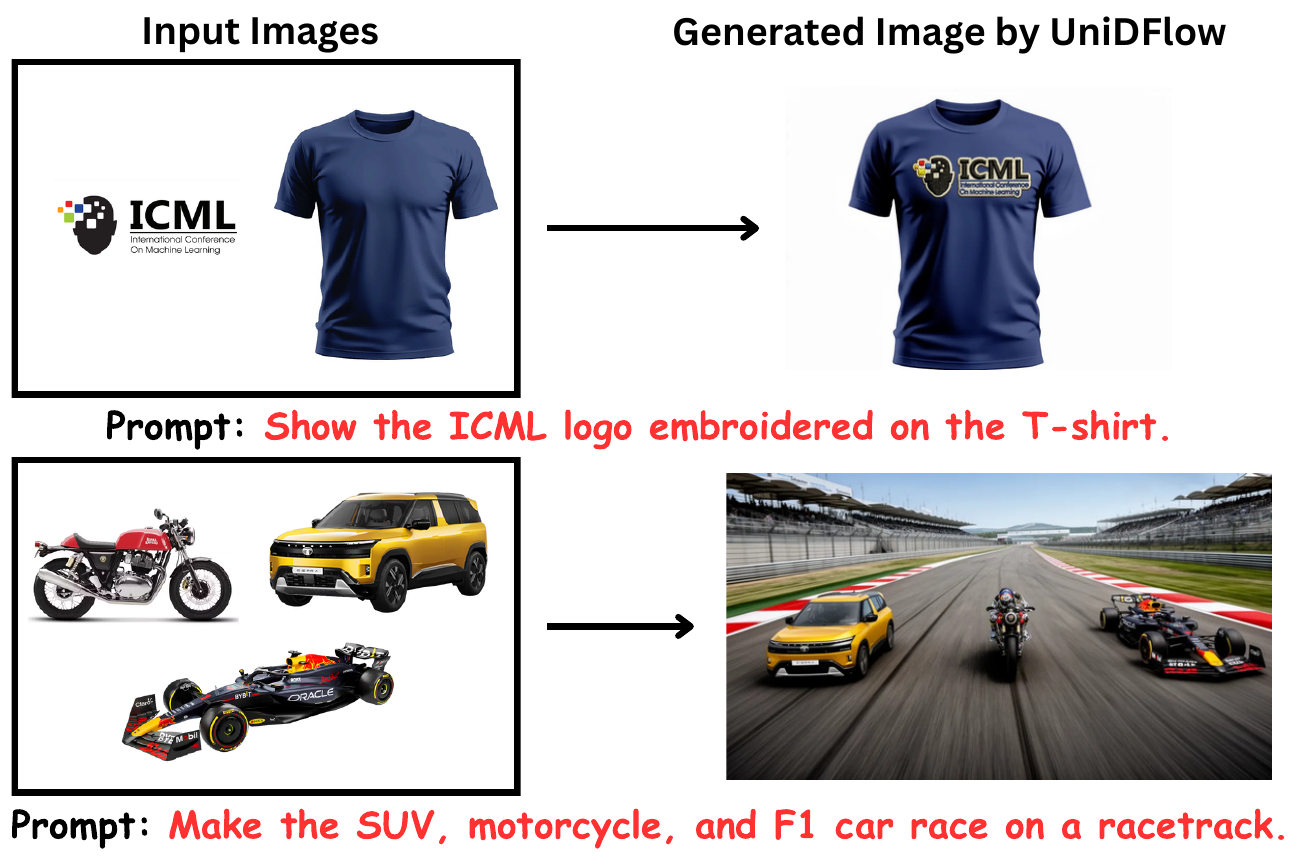}
 \caption{Subject-driven image generation with attribute
editing and multi-object compositional reasoning.}
    \label{fig:multi_object_gen}
\end{figure}}

\newcommand{\Figunderstanding}
{\begin{figure}[t!]
\centering\includegraphics[width=.99\columnwidth]{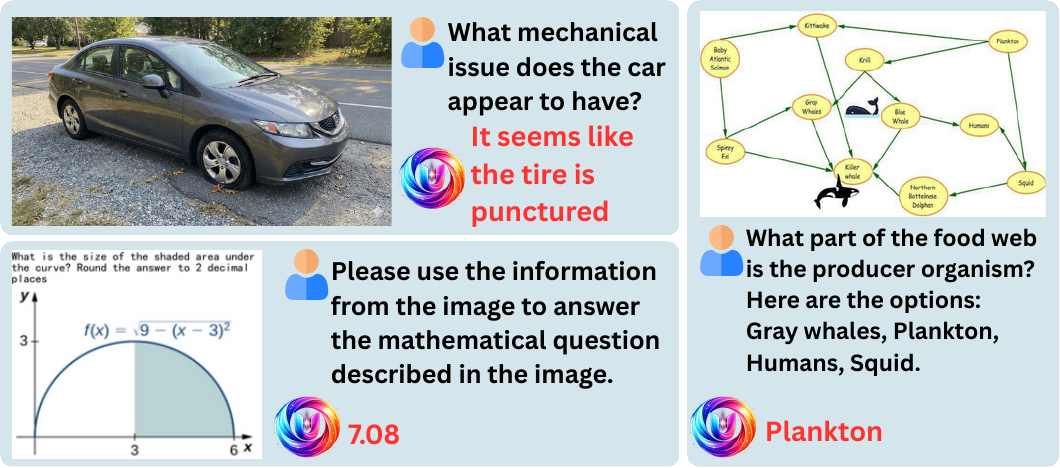}
 \caption{Multimodal reasoning from \modelnamenc }
    \label{fig:img_und1}
\end{figure}}
\newcommand{\TabUnderstanding}{\begin{table*}[t!]
\centering
\small
\caption{Comparison of multimodal understanding performance on
\textsc{EvalVLMBench}~\cite{fu2023mme, liu2024mmbench, yue2024mmmu, yu2024mmvet, lu2024mathvista, tong2024eyes} across diverse reasoning tasks.}
\label{tab:understanding_benchmark}
\setlength{\tabcolsep}{4pt}
\resizebox{\linewidth}{!}{
\begin{tabular}{l l c c c c c c c}
\toprule
\textbf{Model} & \textbf{Params} & \textbf{\textsc{MME-P}} & \textbf{\textsc{MME-S}} & \textbf{\textsc{MMBench}} & \textbf{\textsc{MMMU}} & \textbf{\textsc{MM-Vet}} & \textbf{\textsc{MathVista}} & \textbf{\textsc{MMVP}} \\
\midrule
Qwen2.5-VL~\cite{bai2025qwen2}  & 3B & -- & 2157 & 79.1 & 53.1 & 61.8 & 62.3 & -- \\
BLIP-3~\cite{xue2024xgenmm}  & 4B & -- & -- & 76.8 & 41.1 & -- & 39.6 & -- \\

DeepSeek-VL2~\cite{wu2024deepseekvl2}  & 4B & -- & -- & 51.1 & 60.0 & 62.8 & -- & -- \\
Qwen3-VL~\cite{bai2025qwen3vl}  & 4B & -- & -- & 85.1 & 64.1 & 72.5 & -- & -- \\

\midrule

VILA-U~\cite{wu2024vilau}  & 7B & 1336 & -- & 66.6 & 32.2 & 27.7 & -- & 22.0 \\
Chameleon~\cite{chameleon2024mixedmodal}  & 7B & -- & -- & 35.7 & 28.4 & 8.3 & -- & 0.0 \\
Janus-Pro~\cite{chen2025januspro}  & 7B & 1567 & -- & 79.2 & 41.0 & 50.0 & -- & -- \\

TokenFlow-XL~\cite{geyer2023tokenflow}  & 13B & 1546 & -- & 68.9 & 38.7 & 40.7 & -- & -- \\

BAGEL~\cite{deng2025emerging}  & 7B & 1687 & 2388 & 85.0 & 55.3 & 67.2 & 73.1 & 69.3 \\
OmniGenV2~\cite{wu2025omnigen2}  & 8B & -- & -- & 53.1 & 61.5 & -- & -- & -- \\

EMMA~\cite{he2025emma}  & 4B & -- & -- & 85.8 & 65.1 & 73.0 & 75.8 & -- \\

MammothModa-2~\cite{shen2025mammothmoda2} & 4B & 1753 & 1998 & 86.6 & 71.23 & 79.4 & 81.8 & 77.5 \\ 
MIDDIT~\cite{shi2025muddit} & 4B & 1700   & 1832 &  82.8 & 66.6  &  76.2 & 79.1 &  74.1  \\

\midrule
\textbf{\modelnamenc} & 4B & \textbf{1803} & \textbf{2555} & \textbf{91.2} & \textbf{74.3} & \textbf{82.7} & \textbf{85.9} & \textbf{80.2} \\
\bottomrule
\end{tabular}
}
\vspace{-0.3cm}
\end{table*}}

\newcommand{\TabCombinedGenBench}{
\begin{table}[t!]
\centering
\small
\caption{
Overall generation performance on \textsc{GenEval}~\cite{ghosh2023geneval} and \textsc{DPGBench}~\cite{hu2024ella}. 
Full benchmark-wise breakdowns are provided in Appendix~\ref{sec:full_quant_results}.
}
\label{tab:combined_gen}
\setlength{\tabcolsep}{6pt}
\resizebox{\columnwidth}{!}{
\begin{tabular}{l l  c c}
\toprule
\textbf{Model} & \textbf{Params} & \textbf{GenEval} & \textbf{DPGBench} \\
\midrule

DALL-E 3~\cite{betker2023improving} & -- & 0.67 & 83.50 \\

SD3-Medium~\cite{esser2024scaling} & 2B & 0.74 & 80.43 \\

Qwen-Image(-RL)~\cite{wu2025qwen} & 7B+20B & 0.91 & 88.32 \\

TokenFlow-XL~\cite{geyer2023tokenflow} & 14B & 0.55 & -- \\

Janus-Pro-7B~\cite{chen2025januspro} & 7B & 0.80 & 84.19 \\

BAGEL~\cite{deng2025emerging} & 7B+7B & 0.88 & 87.74 \\

OmniGenV2~\cite{wu2025omnigen2} & 3B+4B & 0.78 & 83.57 \\

MammothModa-2~\cite{shen2025mammothmoda2} & 8B+3B+2B & 0.87 & 87.20 \\

EMMA~\cite{he2025emma} & 4B & 0.93 & 85.63 \\

MUDDIT~\cite{shi2025muddit} & 8B & 0.90 & 86.37 \\

\midrule
\textbf{\modelnamenc} & 4B & \textbf{0.95} & \textbf{91.19} \\
\bottomrule
\end{tabular}
}
\end{table}
}

\newcommand{\TabGenerationEval}{\begin{table*}[t!]
\centering
\small
\setlength{\tabcolsep}{6pt}
\begin{tabular}{l c c c c c}
\toprule
\textbf{Model} & \textbf{Params} &
\textbf{Two Obj} &
\textbf{Counting} &
\textbf{Position} &
\textbf{Overall} \\
\midrule

DALL-E 3~\cite{betker2023improving} & -- & 0.87 & 0.47 & 0.43 & 0.67 \\
SD3-Medium~\cite{esser2024scaling} & 2B & 0.94 & 0.72 & 0.33 & 0.74 \\
Qwen-Image-RL~\cite{wu2025qwen} & 7B+20B & 0.95 & 0.93 & 0.87 & 0.91 \\
TokenFlow-XL~\cite{geyer2023tokenflow} & 14B & 0.60 & 0.41 & 0.16 & 0.55 \\
Janus-Pro-7B~\cite{chen2025januspro} & 7B & 0.89 & 0.59 & 0.79 & 0.80 \\
Bagel~\cite{deng2025emerging} & 7B+7B & 0.95 & 0.84 & 0.78 & 0.88 \\
OmniGenV2~\cite{wu2025omnigen2} & 3B+4B & 0.93 & 0.64 & 0.73 & 0.78 \\
MammothModa-2~\cite{shen2025mammothmoda2} & 8B+3B+2B & 0.97 & 0.63 & 0.90 & 0.87 \\
EMMA~\cite{he2025emma} & 4B & 0.99 & 0.87 & 0.86 & 0.93 \\
MUDDIT~\cite{shi2025muddit} & -- & 0.93 & 0.85 & 0.82 & 0.90 \\
\midrule
\textbf{\modelnamenc} & 4B &
\textbf{0.99} & \textbf{0.89} & \textbf{0.97} & \textbf{0.95} \\
\bottomrule
\end{tabular}

\caption{
Main compositional text-to-image generation results.
We report representative metrics capturing object compositionality (Two Obj),
numerical reasoning (Counting), and spatial understanding (Position).
Full results are provided in Appendix~\ref{sec:appendix_full_results}.
}
\label{tab:genEval_main}
\vspace{-0.4cm}
\end{table*}
}

\newcommand{\TabDPBench}{\begin{table*}[t]
\centering
\small
\setlength{\tabcolsep}{6pt}
\begin{tabular}{l c c c c}
\toprule
\textbf{Model} & \textbf{Params} &
\textbf{Global} &
\textbf{Attribute} &
\textbf{Overall} \\
\midrule

DALL-E 3~\cite{betker2023improving} & -- & 90.97 & 89.39 & 83.50 \\
FLUX.1 [Dev]~\cite{labs2025flux1kontext} & 12B & 74.35 & 88.70 & 83.84 \\
SD3-Medium~\cite{esser2024scaling} & 2B & 87.90 & 88.48 & 80.43 \\
Qwen-Image~\cite{wu2025qwen} & 7B+20B & 91.32 & 92.02 & 88.32 \\

Janus-Pro-7B~\cite{chen2025januspro} & 7B & 86.90 & 89.40 & 84.19 \\
BAGEL~\cite{deng2025emerging} & 7B + 7B & 88.03 & 92.07 & 87.74 \\
OmniGenV2~\cite{wu2025omnigen2} & 3B+4B & 87.98 & 86.43 & 83.57 \\

MammothModa2~\cite{shen2025mammothmoda2} & 8B+3B+2B & 81.16 & 90.16 & 87.20 \\
EMMA~\cite{he2025emma} & 4B & 91.24 & 90.59 & 85.63 \\
\midrule
\textbf{\modelnamenc} & 4B &
\textbf{91.22} & \textbf{95.34} & \textbf{91.19} \\
\bottomrule
\end{tabular}

\caption{
Main fine-grained understanding results.
We report Global scene understanding, Attribute binding performance,
and Overall score. Full category-wise results are provided in
Appendix~\ref{sec:appendix_full_results}\il{fix}.
}
\label{tab:DP_main}
\end{table*}
}

\newcommand{\TabImgEdit}
{\begin{table}[htbp]
\centering
\setlength{\tabcolsep}{3pt}
\resizebox{0.99\columnwidth}{!}{
\begin{tabular}{l c c c c c c}
\toprule
\textbf{Method} 
& \multicolumn{3}{c}{\textbf{\textsc{Emu-Edit}}} 
& \multicolumn{3}{c}{\textbf{\textsc{GEdit-Bench-EN}}} \\
\cmidrule(lr){2-4} \cmidrule(lr){5-7}
& \textbf{CLIP-I}$\uparrow$ 
& \textbf{CLIP-Out}$\uparrow$ 
& \textbf{DINO}$\uparrow$ 
& \textbf{SC}$\uparrow$ 
& \textbf{PQ}$\uparrow$ 
& \textbf{O}$\uparrow$ \\
\midrule
Qwen-Image~\cite{wu2025qwen} & - & - & - & - & - & - \\
FLUX.1 Kontext-Pro~\cite{labs2025flux1kontext} & - & - & - & - & - & - \\
BAGEL & 0.839 & 0.307 & 0.753 & 7.36 & 6.83 & 6.52 \\
UniWorld-V1 & -- & -- & -- & 4.93 & 7.43 & 4.85 \\
OmniGenV2 & 0.876 & 0.309 & 0.822 & 7.16 & 6.77 & 6.41 \\
Emma & 0.911 & 0.311 & 0.834 & 7.33 & 7.54 & 6.52 \\
MammothModa2 & 0.891 & 0.322 & 0.844 & 7.77 & 7.32 & 6.82 \\
\midrule
\textbf{\modelnamenc} 
& \textbf{0.921} 
& \textbf{0.362} 
& \textbf{0.862} 
& \textbf{8.01} 
& \textbf{7.82} 
& 7.12 \\
\bottomrule
\end{tabular}
}
\caption{ Quantitative results on Emu-Edit and GEdit-Bench-EN. Emu-Edit uses CLIP-I/DINO for source consistency and CLIP-Out for caption alignment; GEdit-Bench reports SC (instruction following) and PQ (perceptual quality). }
\label{tab:image_edit}
\vspace{-0.6cm}
\end{table}}

\newcommand{\TabImgEditCombined}{
\begin{table*}[t!]
\centering
\small
\caption{Text-to-image editing results. \textsc{ImgEdit} metric are category-wise scores, while \textsc{Emu-Edit} CLIP-I/DINO are used for source consistency and CLIP-Out for caption alignment. \textsc{GEdit-Bench-EN} evaluates SC (instruction following) and PQ (perceptual quality). 
}
\label{tab:image_edit_combined}
\setlength{\tabcolsep}{4pt}
\resizebox{0.95\linewidth}{!}{
\begin{tabular}{l cccc ccc ccc}
\toprule
\multirow{2}{*}{\textbf{Model}} 
& \multicolumn{4}{c}{\textbf{\textsc{ImgEdit}}} 
& \multicolumn{3}{c}{\textbf{\textsc{Emu-Edit}}}
& \multicolumn{3}{c}{\textbf{\textsc{GEdit-Bench-EN}}} \\

\cmidrule(lr){2-5} \cmidrule(lr){6-8} \cmidrule(lr){9-11}
& \textbf{Add} $\uparrow$ & \textbf{Extract} $\uparrow$  & \textbf{Remove}$\uparrow$  & \textbf{Overall} $\uparrow$ 
& \textbf{CLIP-I} $\uparrow$ & \textbf{CLIP-Out} $\uparrow$ & \textbf{DINO} $\uparrow$ 
& \textbf{SC} $\uparrow$  & \textbf{PQ} $\uparrow$  & \textbf{Overall} $\uparrow$  \\
\midrule

FLUX.1 Kontext-Pro~\cite{labs2025flux1kontext} 
& 4.25 & 2.35 & 3.57 & 4.00 & 0.88 & - & 0.808 & 7.77 & 7.12 & 6.95 \\

BAGEL~\cite{deng2025emerging} 
& 3.56 & 1.70 & 2.62 & 3.20 & 0.839 & 0.307 & 0.753 & 7.36 & 6.83 & 6.52 \\

UniWorld-v1~\cite{lin2025uniworld} 
& 3.82 & 2.27 & 3.24 & 3.26 & -- & -- & -- & 4.93 & 7.43 & 4.85 \\

OmniGenV2~\cite{wu2025omnigen2} 
& 3.57 & 1.77 & 3.20 & 3.44 & 0.876 & 0.309 & 0.822 & 7.16 & 6.77 & 6.41 \\

Emma~\cite{he2025emma} 
& 4.52 & 3.54 & 4.21 & 4.01 & 0.911 & 0.311 & 0.834 & 7.33 & 7.54 & 6.52 \\

MammothModa2~\cite{shen2025mammothmoda2} 
& 4.57 & 3.38 & 3.34 & 4.06 & 0.891 & 0.322 & 0.844 & 7.77 & 7.32 & 6.82 \\

\midrule
\textbf{\modelnamenc} 
& \textbf{4.66} & \textbf{4.01} & \textbf{4.24} & \textbf{4.24}
& \textbf{0.921} & \textbf{0.362} & \textbf{0.862}
& \textbf{8.01} & \textbf{7.82} & \textbf{7.12} \\
\bottomrule
\end{tabular}
}
\vspace{-0.3cm}
\end{table*}
}

\newcommand{\TabImgEditBench}
{\begin{table*}[t!]
\centering
\small
\setlength{\tabcolsep}{5pt}
\begin{tabular}{l c c c c}
\toprule
\textbf{Model} & \textbf{Add} & \textbf{Extract} & \textbf{Remove} & \textbf{Overall} \\
\midrule

Qwen-Image~\cite{wu2025qwen} & 4.38 & 3.43 & 4.14 & 4.27 \\
FLUX.1 Kontext-Pro~\cite{labs2025flux1kontext} & 4.25 & 2.35 & 3.57 & 4.00 \\

Bagel~\cite{deng2025emerging} & 3.56 & 1.70 & 2.62 & 3.20 \\
UniWorld-v1~\cite{lin2025uniworld} & 3.82 & 2.27 & 3.24 & 3.26 \\
OmniGenV2~\cite{wu2025omnigen2} & 3.57 & 1.77 & 3.20 & 3.44 \\

MammothModa2~\cite{shen2025mammothmoda2} & 4.57 & 3.38 & 3.34 & 4.06 \\
Emma~\cite{he2025emma} & 4.52 & 3.54 & 4.21 & 4.01 \\
\midrule
\textbf{Ours} & \textbf{4.66} & \textbf{4.01} & \textbf{4.24} & \textbf{4.24} \\
\bottomrule
\end{tabular}
\caption{Evaluation on key image editing categories. Higher scores indicate better perceived quality and instruction faithfulness.}
\label{tab:image_edit_bench}
\vspace{-0.5cm}
\end{table*}}

\newcommand{\TableAblationDVLM}
{\begin{table}[t!]
\centering
\caption{Ablations. All results are averaged across benchmarks.}
\label{tab:albtaions}
\setlength{\tabcolsep}{1pt}
\resizebox{0.99\columnwidth}{!}{
\begin{tabular}{ccccc}
\toprule
 & \textbf{\textsc{EvalVLM}} & \textbf{\textsc{GenEval}} & \textbf{\textsc{DPGBench}}   & \textbf{\textsc{ImgEdit}}  \\
\midrule
\addlinespace[4pt]
\rowcolor{purple!10}
\textbf{\modelnamenc} & \textbf{82.85} & \textbf{0.95} & \textbf{91.19} & \textbf{4.24} \\
\rowcolor{gray!10}
\multicolumn{5}{c}{\textbf{1. Model Size Ablation}} \\
\addlinespace[2pt]
Qwen3-0.6B  & 79.48 & 0.93 & 88.32 & 4.19 \\
Qwen3-4B   & 82.85 & 0.95 & 91.91 & 4.24 \\
Qwen3-8B   & 84.02 & 0.96 & 92.56 & 4.26 \\
Qwen3-14B  & 89.24 & 0.98 & 95.44 & 4.63 \\
\addlinespace[3pt]
\rowcolor{gray!10}
\multicolumn{5}{c}{\textbf{2. Visual tokenizer}} \\
\addlinespace[2pt]
3D-MBQ-VAE & 81.27 & 0.92 & 91.43 & 4.19 \\
MAGVIT-v2             & 81.19 & 0.91 & 90.34 & 4.16 \\
SweetTok          & 80.76 & 0.92 & 90.44 & 4.12 \\
\addlinespace[3pt]
\rowcolor{gray!10}
\multicolumn{5}{c}{\textbf{3. Architectural Ablations}} \\
\addlinespace[2pt]
w/o LoRA$_{\text{text}}$& 80.11 & 0.92 & 89.33 & 4.01 \\
w/o LoRA$_{\text{img}}$ & 81.23 & 0.93 & 90.05 & 4.08 \\
w/o MoRA & 80.67 & 0.93 & 89.88 & 4.11 \\
Single LoRA (Und+Gen) & 79.92 & 0.90 & 89.12 & 4.08 \\
\addlinespace[3pt]
\rowcolor{gray!10}
\multicolumn{5}{c}{\textbf{4. Loss Function Ablations}} \\
\addlinespace[2pt]
w/o $\mathcal{L}_{\mathrm{vRef-DPO}}$ & 80.45 & 0.91 & 88.44 & 4.09 \\
w/o $\mathcal{L}_{\mathrm{tRef-DPO}}$ & 79.45 & 0.91 & 90.32 & 4.18 \\
w/o $\mathcal{L}_{\mathrm{mRef-DPO}}$ & 77.34 & 0.86 & 86.23 & 4.05 \\
w/o Reflection & 81.23 & 0.89 & 87.57 & 4.14 \\
\addlinespace[3pt]
\rowcolor{gray!10}
\rowcolor{gray!10}
\multicolumn{5}{c}{\textbf{5. Stage-III Alignment Training}} \\
\addlinespace[2pt]
DPO &80.12	&0.92	&88.82	&4.14 \\
uni-GRPO & 80.88	&0.93	&90.07	&4.18\\
\addlinespace[3pt]
\rowcolor{gray!10}
\rowcolor{gray!10}
\multicolumn{5}{c}{\textbf{6. Time Step Conditioning}} \\
\addlinespace[2pt]
In-Context with tokens &82.11	&0.94	&90.09	&4.15 \\
AdLN &81.53	&0.93	&88.75	&4.0 \\

\bottomrule
\end{tabular}
}
\vspace{0.1cm}
\end{table}
}

\newcommand{\TableSceneText}{
\begin{table}
\centering
\caption{Evaluation of existing VLMs and MLLMs on English tasks of OCRBench v2~\cite{fu2024ocrbench} public data. “Recognition”, “Referring”, “Spotting”, “Extraction”, “Parsing”, “Calculation”, “Understanding”, and
“Reasoning” refer to text recognition, text referring, text spotting, relation extraction, element parsing,
mathematical calculation, visual text understanding, and knowledge reasoning, respectively. Higher
values indicate better performance. }
\label{tab:scene_text}
\setlength{\tabcolsep}{6pt}
\resizebox{0.9\linewidth}{!}{
\begin{tabular}{lccccccccc}
\toprule
\textbf{Method} &
\textbf{Recog.} &
\textbf{Ref.} &
\textbf{Spot.} &
\textbf{Extr.} &
\textbf{Pars.} &
\textbf{Calc.} &
\textbf{Und.} &
\textbf{Reas.} &
\textbf{Avg.} \\
\midrule
Qwen2.5-VL-7B~\cite{bai2025qwen2}      & 68.8 & 25.7 & 1.2  & 80.2  & 30.4 & 38.2 & 73.2 & 56.2 & 46.7 \\
InternVL3-14B~\cite{chen2024internvl}     & 67.3 & 36.9 & 11.2 & 89.0  & 38.4 & 38.4 & 79.2 & 60.5 & 52.6 \\
GPT-4o~\cite{openai2024gpt4omini}            & 61.2 & 26.7 & 0.0  & 77.5  & 36.3 & 43.4 & 71.1 & 55.5 & 46.5 \\
GPT-4o-mini~\cite{openai2024gpt4omini}       & 57.9 & 23.3 & 0.6  & 70.8  & 31.5 & 38.8 & 65.9 & 55.1 & 43.0 \\
Gemini-pro~\cite{google_gemini}        & 61.2 & 39.5 & 13.5 & 79.3  & 39.2 & 47.7 & 75.5 & 59.3 & 51.9 \\
Qwen3-VL-8B~\cite{bai2025qwen3vl}       & 64.4 & 38.2 & 5.7  & 91.03 & 37.8 & 44.2 & 76.8 & 62.6 & 55.7 \\
OmniGenV2~\cite{wu2025omnigen2}         & 61.3 & 36.5 & 2.4  & 87.23 & 33.4 & 40.7 & 72.7 & 65.7 & 48.7 \\
BAGEL ~\cite{deng2025emerging}            & 65.8 & 37.1 & 3.3  & 90.45 & 38.5 & 41.3 & 75.2 & 66.4 & 52.2 \\
Emma~\cite{he2025emma}             & 66.7 & 36.5 & 6.7  & 91.3  & 37.5 & 44.5 & 76.7 & 67.2 & 53.8 \\
MUDDIT~\cite{shi2025muddit}            & 64.9 & 38.4 & 13.7 & 92.6  & 34.5 & 49.4 & 78.3 & 66.1 & 54.7 \\
MammothModa2~\cite{shen2025mammothmoda2}      & 68.2 & 39.5 & 11.4 & 92.2  & 39.1 & 50.2 & 80.2 & 68.1 & 56.1 \\
\midrule
\textbf{UniDFlow-4B}  & 69.9 & 41.2 & 12.9 & 94.1  & 42.2 & 53.4 & 83.1 & 70.8 & 58.4 \\
\textbf{UniDFlow-8B}  & 72.2 & 43.8 & 14.9 & 95.0  & 45.7 & 55.1 & 85.9 & 73.5 & 60.7 \\
\textbf{UniDFlow-14B} & \textbf{76.7} & \textbf{47.1} & \textbf{16.5} & \textbf{96.9} & \textbf{48.4} & \textbf{58.1} & \textbf{88.7} & \textbf{77.1} & \textbf{63.8} \\
\bottomrule
\end{tabular}
}
\end{table}
}

\newcommand{\TableAblationPyratok}{
\begin{wraptable}{r}{0.7\linewidth}
\centering
\caption{Ablation study on visual tokenizer. All results are averaged across benchmarks.}
\small
\setlength{\tabcolsep}{3pt}
\begin{tabular}{lcccc}
\toprule
 & Text Gen $\uparrow$ & GenEval $\uparrow$ & DPGBench $\uparrow$ & ImgEdit $\uparrow$ \\
\midrule
3D-MBQ-VAE~\cite{susladkar2024motionaura} & 81.27 & 0.92 & 91.43 & 4.19 \\
MAGVIT-v2~\cite{luo2024open}             & 81.19 & 0.91 & 90.34 & 4.16 \\
SweetTok~\cite{tan2025sweettok}          & 80.76 & 0.92 & 90.44 & 4.12 \\
PyraTok~\cite{susladkar2026pyratok}  & \textbf{82.85} & \textbf{0.95} & \textbf{91.91} & \textbf{4.24} \\ 
\bottomrule
\end{tabular}
\vspace{-6pt}
\label{tab:tokenizer}
\end{wraptable}
}

\newcommand{\TableSuppGen}{
\begin{table}[t!]
\centering
\small
\caption{Detailed text-to-image generation evaluation on \textsc{DPG-Bench} and \textsc{GenEval}.}
\label{tab:t2i_supp}
\setlength{\tabcolsep}{5.2pt}
\renewcommand{\arraystretch}{1.05}
\resizebox{0.9\linewidth}{!}{
\begin{tabular}{l c c c c c c c}
\toprule

\rowcolor{gray!18}
\multicolumn{8}{c}{\textbf{\textsc{DPG-Bench}}} \\
\midrule
\textbf{Model} & \textbf{Params} & \textbf{Global} & \textbf{Entity} & \textbf{Attribute} & \textbf{Relation} & \multicolumn{2}{c}{\textbf{Other}} \\
\midrule
DALL-E 3~\cite{betker2023improving}                  & --          & 90.97 & 89.61 & 89.39 & 90.58 & \multicolumn{2}{c}{89.83} \\
SD3-Medium~\cite{esser2024scaling}                   & 2B          & 87.92 & 91.01 & 88.48 & 80.72 & \multicolumn{2}{c}{86.81} \\
Qwen-Image~\cite{wu2025qwen}     & 7B+20B      & 91.32 & 91.56 & 92.02 & 94.31 & \multicolumn{2}{c}{92.73} \\
TokenFlow-XL~\cite{geyer2023tokenflow}               & 14B         & 87.33 & 88.54 & 89.01 & 85.09 & \multicolumn{2}{c}{86.55} \\
Janus-Pro-7B~\cite{chen2025januspro}                 & 7B          & 86.91 & 88.95 & 89.43 & 90.02 & \multicolumn{2}{c}{89.48} \\
BAGEL~\cite{deng2025emerging}                        & 7B+7B       & 89.42 & 91.43 & 90.42 & 92.34 & \multicolumn{2}{c}{88.78} \\
OmniGenV2~\cite{wu2025omnigen2}                       & 3B+4B       & --    & --    & 86.43 & 91.23 & \multicolumn{2}{c}{--} \\
MammothModa-2~\cite{shen2025mammothmoda2}            & 8B+3B+2B    & 81.16 & 92.99 & 90.16 & 94.35 & \multicolumn{2}{c}{84.81} \\
EMMA~\cite{he2025emma}                               & 4B          & 91.24 & 91.71 & 90.59 & 92.23 & \multicolumn{2}{c}{90.02} \\
MUDDIT~\cite{shi2025muddit}                          & 7B          & 89.42 & 90.47 & 89.56 & 90.72 & \multicolumn{2}{c}{88.63} \\
\midrule
\textbf{UniDFlow}                                    & 4B          & \textbf{93.42} & \textbf{94.44} & \textbf{95.34} & \textbf{95.03} & \multicolumn{2}{c}{\textbf{93.86}} \\

\midrule
\rowcolor{gray!18}
\multicolumn{8}{c}{\textbf{\textsc{GenEval}}} \\
\midrule
\textbf{Model} & \textbf{Params} & \textbf{Single Obj} & \textbf{Two Obj} & \textbf{Counting} & \textbf{Colors} & \textbf{Position} & \textbf{Color Attr} \\
\midrule
DALL-E 3~\cite{betker2023improving}                  & --        & 0.96 & 0.87 & 0.47 & 0.83 & 0.43 & 0.45 \\
SD3-Medium~\cite{esser2024scaling}                   & 2B        & 0.99 & 0.94 & 0.72 & 0.89 & 0.33 & 0.60 \\
Qwen-Image~\cite{wu2025qwen}     & 7B+20B    & \textbf{1.00} & 0.95 & \textbf{0.93} & 0.92 & 0.87 & 0.83 \\
TokenFlow-XL~\cite{geyer2023tokenflow}               & 14B       & 0.95 & 0.60 & 0.41 & 0.81 & 0.16 & 0.24 \\
Janus-Pro-7B~\cite{chen2025januspro}                 & 7B        & 0.99 & 0.89 & 0.59 & 0.90 & 0.79 & 0.66 \\
BAGEL~\cite{deng2025emerging}                        & 7B+7B     & 0.98 & 0.95 & 0.84 & 0.95 & 0.78 & 0.77 \\
OmniGenV2~\cite{wu2025omnigen2}                       & 3B+4B     & 0.95 & 0.93 & 0.64 & 0.81 & 0.73 & 0.74 \\
MammothModa-2~\cite{shen2025mammothmoda2}            & 8B+3B+2B  & \textbf{1.00} & 0.97 & 0.63 & 0.89 & \textbf{0.90} & 0.82 \\
EMMA~\cite{he2025emma}                               & 4B        & \textbf{1.00} & \textbf{0.99} & 0.87 & \textbf{0.98} & 0.86 & \textbf{0.87} \\
MUDDIT~\cite{shi2025muddit}                          & 7B        & 0.95 & 0.93 & 0.85 & 0.96 & 0.82 & 0.84 \\
\midrule
\textbf{UniDFlow}                                    & 4B        & \textbf{1.00} & \textbf{0.99} & 0.89 & \textbf{0.98} & 0.87 & 0.83 \\
\bottomrule
\end{tabular}
}
\end{table}
}

\newcommand{\tableeditcomplexity}{
\begin{table}[t]
\centering
\caption{Analysis of the reflection mechanism across editing tasks with varying complexity.}
\label{tab:reflection_editing}
\setlength{\tabcolsep}{3pt}
\renewcommand{\arraystretch}{1.1}
\resizebox{\columnwidth}{!}{%
\begin{tabular}{llccccc}
\toprule
\textbf{Setting} & \textbf{Type} & \textbf{PSNR}$\uparrow$ & \textbf{SSIM}$\uparrow$ & \textbf{SD}$\downarrow$ & \textbf{CLIP$_{\text{ed}}$}$\uparrow$ & \textbf{CLIP$_{\text{un}}$}$\uparrow$ \\
\midrule
\multirow{4}{*}{Refl.}
 & Add/Rem.   & 31.43          & 0.856          & \phantom{0}9.12         & 26.67          & 25.05          \\
 & Reason.    & 27.82          & 0.869          & \phantom{0}\textbf{7.45} & 27.53          & 24.11          \\
 & Sc. Text   & 30.11          & \textbf{0.891} & 10.43                   & \textbf{31.44} & \textbf{27.62} \\
 & Multi-Obj. & \textbf{31.45} & 0.847          & \phantom{0}9.52         & 28.82          & 25.01          \\
\midrule
\multirow{4}{*}{w/o Refl.}
 & Add/Rem.   & 29.44 & 0.833 & 11.33 & 24.44 & 23.37 \\
 & Reason.    & 23.67 & 0.849 & 10.02 & 23.12 & 23.78 \\
 & Sc. Text   & 27.52 & 0.871 & 12.02 & 27.72 & 25.28 \\
 & Multi-Obj. & 28.89 & 0.825 & 11.78 & 25.17 & 21.75 \\
\bottomrule
\end{tabular}%
}
\end{table}
}
\begin{abstract}
We propose \modelnamenc{}, a unified discrete flow-matching framework for multimodal understanding, image generation, and instruction-guided editing. \modelnamenc{} decouples understanding and generation via task-specific low-rank adapters, avoiding objective interference and representation entanglement, while a novel reference-based multimodal preference alignment optimizes relative outcomes under identical conditioning, improving faithfulness and controllability without large-scale retraining. 
\modelnamenc{} achieves SOTA performance across six benchmarks and exhibits strong zero-shot generalization to tasks including inpainting, in-context image generation, reference-based editing, and compositional generation, despite no explicit task-specific training.\looseness-1\\
\noindent \logoicon~\href{https://plan-lab.github.io/unidflow}{\textcolor{IllinoisBlue}{PLAN Lab}~\textcolor{IllinoisOrange}{https://plan-lab.github.io/unidflow}}
\end{abstract}

\section{Introduction}
Multimodal generative systems have become central to everyday productivity, with large language models (LLMs) such as ChatGPT and Gemini~\cite{google_gemini} enabling strong reasoning and instruction following.
Similarly, diffusion-based models such as Stable Diffusion~\cite{rombach2022high, esser2024scaling} and DALL·E~\cite{ramesh2021zero, betker2023improving} excel at high-fidelity image and video generation. However, these models remain largely disjoint as LLM-centric models excel at understanding but lack native generative mechanisms, while diffusion models provide powerful generation with limited semantic grounding and reasoning. This separation motivates unified multimodal models that integrate LLM-level understanding with diffusion-level generation within a single architecture~\cite{wang2024multi, xie2024show}.\looseness-1

Early approaches in this direction, such as Emu~\cite{dai2023emu} and Chameleon ~\cite{chameleon2024mixedmodal}, model both text and vision using a single auto-regressive (AR) transformer~\cite{vaswani2017attention}. While simple, AR-based generation is highly inefficient for high-dimensional visual outputs. Hybrid models, including EMMA~\cite{he2025emma}, OmniGenV2~\cite{wu2025omnigen2}, MammothModa2~\cite{shen2025mammothmoda2}, and BAGEL~\cite{deng2025emerging}, combine AR modeling for text with diffusion-style objectives for images to retain language understanding while improving generation. UniDisc~\cite{swerdlow2025unified} and MUDDIT~\cite{shi2025muddit} employ fully discrete diffusion with a unified denoising objective for text and images, but performance lags behind hybrid models. 

Despite recent progress, existing unified models still face several fundamental limitations: \textbf{(1)} Large-scale AR--diffusion frameworks couple cross-entropy decoding with diffusion-style regression~\cite{shen2025mammothmoda2, wu2025omnigen2}, creating mismatched objectives that lead to unstable joint optimization.
\textbf{(2)} Even with strong pretrained initialization, many approaches rely on full-model updates over hundreds of millions of samples~\cite{deng2025emerging, he2025emma}, incurring substantial compute while often degrading general-purpose reasoning ability.
\textbf{(3)} Current unified diffusion approaches entangle understanding and generation within shared parameters, thus improving one capability can inadvertently erode the other~\cite{zhong2026unified, shi2025muddit}.
\textbf{(4)} Generation and editing are often improved through additional alignment stages, such as multimodal reflection~\cite{wu2025omnigen2} or reinforcement learning with scalar rewards~\cite{shen2025mammothmoda2}. However, these approaches optimize outputs in isolation, encouraging higher scores or improved reasoning trajectories without modeling relative preference under identical conditioning. As a result, they fail to learn explicit decision boundaries between faithful and subtly incorrect edits.\looseness-1 

To address the aforementioned limitations, we introduce \modelnamenc, a unified discrete diffusion framework for efficient multimodal understanding and generation. \modelnamenc leverages a strong pretrained vision–language model as a prior, avoiding redundant pretraining and enabling parameter-efficient adaptation through lightweight adapters. We perform large-scale three-stage training: (i) an understanding-focused stage, (ii) a generation-focused stage, and (iii) a joint understanding–generation stage with reference-based multimodal preference optimization to improve editing fidelity and controllability. To prevent parameter entanglement, \modelnamenc trains separate adapters for understanding and generation, while the final stage trains only a lightweight router to combine them dynamically.\looseness-1

 Fig.~\ref{fig:activation_map} visualizes the instruction-guided activation maps during editing. \modelnamenc consistently attends more precisely to instruction-relevant regions, whether modifying coarse objects (\eg adding a T-shirt) or finer details (\eg changing the swoosh color).
Our main contributions are:
\begin{itemize}[itemsep=0.2ex, parsep=0pt, topsep=-1pt]
    \item We introduce \modelnamenc, a unified discrete diffusion model that repurposes a pretrained vision--language backbone as a generator over multimodal tokens, enabling understanding, text generation, image synthesis, and editing within one probabilistic interface.\looseness-1
    \item We unify text and image generation under a \emph{single discrete flow-matching objective} for all tasks and incorporate a {stable time-conditioning} mechanism that preserves the backbone's reasoning priors. Compared to prior multi-objectives, \modelnamenc achieves efficient training and inference, requiring only 20 denoising steps while preserving high generation quality.
    \item We propose \dponamenc{}, a \emph{reference-guided multimodal preference alignment} that directly optimizes a pairwise log-likelihood margin between preferred and non-preferred outcomes, anchored to a frozen reference model, leading to more faithful and controllable editing behavior and stabilized training by enforcing comparative preference learning and rewarding improvements over the base model rather than absolute score maximization.
    \item \modelnamenc achieves state-of-the-art performance on eight benchmarks spanning understanding, generation, and editing, with up to 13\% improvement over larger unified models with more than 3$\times$ parameters, and up to 24\% gains over popular models such as Qwen 3~\cite{bai2025qwen3vl} and DeepSeek-VL2~\cite{wu2024deepseekvl2}.

\end{itemize}

\FigIntro
\section{Related Work}

\noindent \textbf{Diffusion for Visual Generation.}
Diffusion probabilistic models (DPMs)~\cite{ho2020denoising,nichol2021glide,saharia2022photorealistic} outperform GANs~\cite{goodfellow2014generative} in stability and quality but are costly in pixel space. Latent diffusion models (LDMs)~\cite{rombach2022high} mitigate this via compressed latent representations, enabling strong text-to-image generation~\cite{zhang2023adding,chen2023pixart,podell2023sdxl}. Discrete diffusion~\cite{austin2021structured} extends diffusion to categorical spaces using masking-based corruption, motivating parallel mask-and-predict generators that improve fidelity and efficiency~\cite{gu2022vector,chang2022maskgit}.\looseness-1

\noindent \textbf{Unified Models for Understanding and Generation.}
To unify understanding and generation, early works such as Emu~\cite{dai2023emu,sun2024generative} and Chameleon~\cite{chameleon2024mixedmodal} adopt fully autoregressive modeling over text and visual tokens, but scale poorly for high-resolution images. Hybrid frameworks including EMMA~\cite{he2025emma}, OmniGenV2~\cite{wu2025omnigen2}, MammothModa2~\cite{shen2025mammothmoda2}, and BAGEL~\cite{deng2025emerging} combine autoregressive text modeling with diffusion-based image generation, yet still face modality and objective mismatches. Fully discrete diffusion models such as UniDisc~\cite{swerdlow2025unified} and MUDDIT~\cite{shi2025muddit} further unify modeling but lag behind large-scale hybrids. Our work introduces \modelnamenc{}, a unified discrete flow-matching model with stable time-conditioning that preserves reasoning priors and enables efficient, high-fidelity multimodal generation and editing.

\noindent \textbf{LLM and Diffusion Preference Alignment.}
LLMs~\cite{touvron2023llama,liu2024deepseek} provide strong reasoning with autoregressive Transformers, and VLMs~\cite{bai2023qwen,bai2025qwen2} extend them to images by projecting visual features (\eg SigLIP~\cite{zhai2023sigmoid}) into the language token space. 
Qwen~\cite{bai2025qwen3vl}, LLaVA~\cite{liu2023visual}, BLIP-2~\cite{li2023blip}, and Flamingo~\cite{alayrac2022flamingo} excel at multimodal understanding but typically rely on separate diffusion backbones for image generation and editing. Preference learning has also been adapted to diffusion models, including Diffusion-DPO~\cite{wallace2024diffusion}, score-space alignment (DSPO)~\cite{zhu2025dspo}, and stabilized variants such as DGPO~\cite{luo2025reinforcing} and discrete-diffusion extensions~\cite{borso2025preference}. Prior work further improves controllability via additional alignment stages (\eg multimodal reflection~\cite{wu2025omnigen2} or scalar-reward RL~\cite{shen2025mammothmoda2}). In contrast, \modelnamenc performs \emph{reference-based multimodal preference alignment}, optimizing a pairwise log-likelihood margin against a frozen reference model for stable, comparative supervision, improving faithfulness and controllable editing.

\section{Method}
\subsection{Preliminaries: Discrete Flow Matching}
We use  Discrete Flow Matching (DFM)~\cite{gat2024discrete} as the common objective across all training stages. DFM learns a transport field in discrete spaces by mapping samples from noise to data. Let $x_0 \sim q_{\text{data}}$ denote a clean discrete sample (\eg text or visual tokens), and $x_t$ its corrupted version at time step $t \in \{0,\dots,T\}$ generated by a fixed forward noising process $q(x_t \mid x_0, t)$.
Given $x_t$, a flow network $f_\theta(x_t, t, c)$ conditioned on time $t$ and context $c$ predicts the transport toward the clean state as $
f_\theta(x_t, t, c) \approx q(x_0 \mid x_t, t, c)$.
The model is trained by minimizing a token-wise categorical negative log-likelihood:
\begin{equation}
\setlength{\abovedisplayskip}{6pt}
\setlength{\belowdisplayskip}{6pt}
\scalebox{0.9}{$
\mathcal{L}_{\text{DFM}}(\theta; x_0 \mid x_t, t, c)
=
\mathbb{E}_{x_0,t,x_t}
\left[
- \log f_\theta(x_0 \mid x_t, t, c)
\right].
$}
\label{eq:dfm_loss}
\end{equation}
At inference, sampling starts from $x_T \sim q_{\text{noise}}$ and applies the learned flow to recover $x_0$. By directly estimating transport directions, DFM enables efficient few-step sampling, with conditioning via $c$ supporting unified language modeling, visual generation, and editing. 

\subsection{\modelnamenc}
We cast multimodal understanding, conditional generation, and instruction-based image editing as a single discrete denoising process.
Starting from a pretrained vision--language transformer with parameters $\theta_0$, \modelnamenc learns to recover a clean token sequence from a corrupted one under appropriate conditioning.
For understanding, the denoised sequence corresponds to answer text tokens conditioned on an instruction $p$ and an input image $x$; for generation and editing, it corresponds to visual tokens conditioned on $p$ and a reference image $x_{\mathrm{ref}}$.
To enable discrete diffusion over images, we map images to sequences of discrete visual tokens using a pretrained tokenizer, and we use bidirectional self-attention to support full-context denoising.
All task-specific adaptation is implemented with low-rank adapters (LoRA), while $\theta_0$ remains frozen. 
\FigArch

\FigArchStageThree
Our training follows a three-stage pipeline (illustrated in Figs.~\ref{fig:overview_stage12} and~\ref{fig:overview_stage3}): Stage~I aligns the pretrained vision–language backbone for diffusion-based multimodal understanding, Stage~II adapts the model for discrete visual generation while preserving reasoning capabilities, and Stage~III performs reference-based multimodal preference alignment to improve fidelity and controllability. We first describe the time-conditioned normalization used throughout the model, followed by the three training stages.\looseness-1

\subsubsection{Time-Step Guided RMSNorm}
Conditioning a pretrained transformer on diffusion time by directly adding time embeddings to attention or MLP activations can destabilize training by perturbing learned feature distributions. We address this with Time-Step Guided RMSNorm (TSG-RMSNorm), which injects time information by modulating the RMSNorm scale parameters rather than altering the activations themselves. This preserves pretrained representations by keeping the direction of hidden states unchanged while only applying a controlled, time-dependent rescaling.

Let $h_{\ell}\in\mathbb{R}^d$ denote the input hidden state (activation vector) to the RMSNorm layer at transformer layer $\ell$.
Standard RMSNorm is $\mathrm{RMSNorm}(h_{\ell})\!=\!\gamma_\ell \odot \frac{h_{\ell}}{\mathrm{RMS}(h_{\ell})}$, where $\mathrm{RMS}(h_{\ell})=\sqrt{\tfrac{1}{d}\sum_{j=1}^d h_{\ell,j}^2+\varepsilon}.$
Given a time embedding $e(t)$, we predict a time-dependent modulation for each layer, \ie
$s_\ell(t)=W^{(s)}_\ell e(t), ~~ b_\ell(t)=W^{(b)}_\ell e(t)$. We apply these to the pretrained RMSNorm parameters via
\begin{equation}
\setlength{\abovedisplayskip}{6pt}
\setlength{\belowdisplayskip}{6pt}
\begin{aligned}
\mathrm{TSG\mbox{-}RMSNorm}(h_{\ell},t)\!=\! \mathrm{RMSNorm}(h_{\ell})\\\odot\big(\gamma_\ell\odot(1+s_\ell(t))\big)+b_\ell(t),
\end{aligned}
\end{equation}
where $\gamma_\ell$ is the pretrained RMSNorm scale and $\odot$ denotes element-wise multiplication.
All time-modulation parameters are zero-initialized so that $s_\ell(t)=0$ and $b_\ell(t)=0$ at initialization, exactly recovering the pretrained model.

\subsubsection{Stage I: Text Alignment }
Unified multimodal models often entangle understanding and generation objectives, leading to representational interference and degraded reasoning.
We first adapt the pretrained backbone to diffusion-style understanding through text alignment in isolation, preserving language--visual reasoning before introducing generative training.

Given an instruction $p$, visual tokens $x$, and a fully masked text token sequence $y_{\mathrm{txt,t}}$, the model predicts the clean answer tokens $y_{\mathrm{txt,0}}$ using discrete flow matching. The training objective follows Eq.(~\ref{eq:dfm_loss}):
\begin{equation}
\setlength{\abovedisplayskip}{4pt}
\setlength{\belowdisplayskip}{4pt}
\mathcal{L}_{\mathrm{under}}
=
\mathcal{L}_{\mathrm{DFM}}\!\big(\Delta\theta_{u}\,;\, y_{\mathrm{txt},0} \mid y_{\mathrm{txt},t},\, p,\, x\big).
\end{equation}

where $\theta_0$ denotes frozen pretrained VLM parameters and $\Delta\theta_u$ are \(LoRA_{text}\) adapters specialized for understanding.

To prevent semantic drift from the pretrained language behavior, we additionally regularize the diffusion-predicted distribution with a KL divergence against the autoregressive answer distribution produced by the original VLM:
\begin{equation}
\setlength{\abovedisplayskip}{4pt}
\setlength{\belowdisplayskip}{4pt}
\scalebox{0.8}{$
\mathcal{L}_{\mathrm{KL}}
=
\mathrm{KL}\!\Big(
p_{\mathrm{DFM}}\!\big(y_{\mathrm{txt},0} \mid y_{\mathrm{txt},t},\, t,\, p,\, x\big)
\;\|\;
p_{\mathrm{AR}}\!\big(y_{\mathrm{txt}} \mid p,\, x\big)
\Big).
$}
\end{equation}
This constraint anchors diffusion-based decoding to the pretrained linguistic manifold while allowing bidirectional attention and time-conditioned normalization to support non-autoregressive reasoning. 

The total Stage I objective is $
\mathcal{L}_{\mathrm{Stage\ I}}\!=\!\mathcal{L}_{\mathrm{under}}\!+\!\lambda_{\mathrm{KL}} \mathcal{L}_{\mathrm{KL}}$.

\subsubsection{Stage II: Vision Alignment}
This stage adapts the same frozen backbone for conditional generation in discrete visual token space, while preserving the understanding behavior learned in the previous training stage.
We keep $\theta_0$ and $\Delta\theta_u$ frozen and introduce a separate set of LoRA adapters $\Delta\theta_g$ specialized for generation.\looseness-1

Given an instruction $p$ and corrupted visual tokens $y_{\mathrm{vis, t}}$, the model predicts clean visual tokens $y_{\mathrm{vis,0}}$ using discrete flow matching with parameters $\theta_0+\Delta\theta_u+\Delta\theta_g$:
\begin{equation}
\setlength{\abovedisplayskip}{6pt}
\setlength{\belowdisplayskip}{6pt}
\mathcal{L}_{\mathrm{Stage\ II}}
=
\mathcal{L}_{\text{DFM}}\!\big(\Delta\theta_g \,;\, y_{\mathrm{vis,0}} \mid y_{\mathrm{vis,t}}, t, p\big),
\end{equation}
where only $\Delta\theta_g$ ( \(LoRA_\text{img}\)) is trainable, while \(\theta_0\) and the understanding adapters \(\Delta\theta_u\) are kept frozen.
The diffusion process operates entirely in a discrete latent space, enabling efficient sampling and seamless integration with the backbone's token-based architecture. By isolating generation-specific parameters, Stage II establishes strong conditional image generation capabilities without interfering with the language and reasoning behavior learned during Stage I.

\subsubsection{Stage III: Reference-Based Multimodal Preference Alignment}
While the previous stages endow \modelnamenc with strong multimodal understanding and generation capabilities, token-level likelihood training cannot reliably distinguish between multiple plausible outputs that differ in instruction fidelity, visual grounding, or reasoning consistency. Stage~III therefore introduces a {reference-based multimodal preference alignment} objective that explicitly optimizes relative preferences across text, vision, and reflection, grounded in reference images.\looseness-1

Each preference instance specifies an instruction $p$ with paired preferred($w$)/rejected($l$) outcomes: reference image ($x_{\mathrm{ref}}^{w}, x_{\mathrm{ref}}^{l}$),
text responses $(y_{\mathrm{txt}}^{w}, y_{\mathrm{txt}}^{l})$,
visual tokens $(y_{\mathrm{vis}}^{w}, y_{\mathrm{vis}}^{l})$,
and reflection sequences $(r^{w}, r^{l})$, allowing the model to learn which multimodal outcomes are preferred, conditioned on both the instruction and the visual reference.

\noindent \textbf{Mixture-of-LoRA Routing (MoRA).}
Since this stage optimizes preferences for both understanding and generation, naively sharing parameters can introduce objective interference, while static routing restricts adaptability. Therefore, we learn a lightweight router $r_\phi$ with parameters $\phi$ that dynamically composes task-specific adapters based on the hidden state at diffusion step $t$: \looseness-1
\begin{equation}
\setlength{\abovedisplayskip}{6pt}
\setlength{\belowdisplayskip}{6pt}
\Delta\theta(t) = \alpha_t \Delta\theta_u + (1-\alpha_t)\Delta\theta_g,
\quad
\alpha_t = r_\phi(h_t),
\end{equation}

\noindent \textbf{Multimodal Preference Learning.}
We adopt a reference-anchored  Direct Preference Optimization (DPO) objective with a frozen reference policy $\pi_{\mathrm{ref}}$.
For text, the loss is 
$\mathcal{L}_{\mathrm{tRef-DPO}}\!=\!- \log \sigma\left(\beta \Delta^{txt}_\theta\right)$ and  preference margin is
\begin{equation}
\setlength{\abovedisplayskip}{6pt}
\setlength{\belowdisplayskip}{6pt}
\scalebox{0.99}{$
\Delta_{\theta}^{\mathrm{txt}}
\!=\!
\log \frac{\pi_\theta(y_{\mathrm{txt}}^{w} \mid p, x_{\mathrm{ref}}^{w})}{\pi_{\mathrm{ref}}(y_{\mathrm{txt}}^{w} \mid p, x_{\mathrm{ref}}^{w})}
\!-\!
\log \frac{\pi_\theta(y_{\mathrm{txt}}^{l} \mid p, x_{\mathrm{ref}}^{l})}{\pi_{\mathrm{ref}}(y_{\mathrm{txt}}^{l} \mid p, x_{\mathrm{ref}}^{l})},
$}
\end{equation}
For vision, we concatenate reflection and image tokens as $\tilde{y}_{\mathrm{vis}}=(r,y_{\mathrm{vis}})$, similarly 
$\mathcal{L}_{\mathrm{vRef-DPO}}\!=\!- \log \sigma\left(\beta \Delta^{vis}_\theta\right)$.
\setlength{\abovedisplayskip}{6pt}
\setlength{\belowdisplayskip}{6pt}
\begin{equation}
\scalebox{0.99}{$
\Delta_{\theta}^{\mathrm{vis}}
=
\log \frac{\pi_\theta(\tilde{y}_{\mathrm{vis}}^{w} \mid p, x_{\mathrm{ref}}^{w})}{\pi_{\mathrm{ref}}(\tilde{y}_{\mathrm{vis}}^{w} \mid p, x_{\mathrm{ref}}^{w})}
-
\log \frac{\pi_\theta(\tilde{y}_{\mathrm{vis}}^{l} \mid p, x_{\mathrm{ref}}^{l})}{\pi_{\mathrm{ref}}(\tilde{y}_{\mathrm{vis}}^{l} \mid p, x_{\mathrm{ref}}^{l})}.
$}
\end{equation}
Stage III jointly aligns text and vision through a preference-augmented objective:
$\mathcal{L}_{\mathrm{mRef\text{-}DPO}} = \lambda_t \mathcal{L}_{\mathrm{tRef\text{-}DPO}} + \lambda_v \mathcal{L}_{\mathrm{vRef\text{-}DPO}},$
which promotes faithful instruction following, grounded visual editing, and consistent multimodal behavior under reference conditioning. 

We optimize a unified objective that combines discrete flow-matching (DFM) likelihood training for three output streams, text generation, $\mathcal{L}_{\text{text}}
=\mathcal{L}_{\text{DFM}}\!\left(\phi;y^{w}_{\mathrm{txt},0}\mid y^{w}_{\mathrm{txt},t},x^{w}_{\text{ref}},p,t\right),$ visual editing $
\mathcal{L}_{\text{edit}}
=\mathcal{L}_{\text{DFM}}\!\left(\phi;y^{w}_{\mathrm{vis},0}\mid y^{w}_{\mathrm{vis},t},x^{w}_{\text{ref}},p,t\right),$ and reflection, with reference-anchored multimodal preference alignment $\mathcal{L}_{\text{refl}}
=\mathcal{L}_{\text{DFM}}\!\left(\phi;r^{w}_0\mid r^{w}_t,x^{w}_{\text{ref}},y^{w}_{\text{vis},t},y^{w}_{\text{txt},t},p,p_{\text{edit}},t\right)$. Thus, the final objective for Stage III is:
\begin{equation}
\setlength{\abovedisplayskip}{4pt}
\setlength{\belowdisplayskip}{4pt}
\mathcal{L}_{\text{Stage-III}}
=\mathcal{L}_{\text{text}}+\mathcal{L}_{\text{edit}}+\mathcal{L}_{\text{refl}}+\mathcal{L}_{\mathrm{mRef\text{-}DPO}}
\end{equation}

The DFM terms maximize time-conditioned token likelihood along the discrete diffusion trajectory under their respective conditionings (instruction, reference image, or edit prompt), enforcing token-level consistency. The $\mathcal{L}_{\mathrm{mRef\text{-}DPO}}$ term introduces comparative alignment by increasing the log-likelihood margin of preferred over rejected outputs relative to a frozen reference policy $\pi_{\mathrm{ref}}$, stabilizing training and improving cross-modal faithfulness.

\section{Experiments}

\TabUnderstanding 

We conduct extensive experiments to evaluate the performance of \modelnamenc across six benchmarks, covering multimodal understanding, generation, and editing.  In Stage I, we train using \textsc{MMInstruct}~\cite{liu2024mminstruct} to establish strong multimodal understanding. Stage II focuses on generative capability by training on \textsc{Text-to-Image-4M}~\cite{jackyhate2024texttoimage2m, sun2023journeydb, schuhmann2022laion5b} . Stage III performs reference-based multimodal preference alignment with 3.5M curated preference samples under identical inputs and reference images. 
Dataset construction for preference alignment, training, and implementation details are provided in Appendices \ref{sec:implementation}-\ref{sec:data}.

\subsection{Multi-Modal Understanding}
Table~\ref{tab:understanding_benchmark} reports results on the \textsc{EvalVLM} benchmark. 
Compared to strong unified hybrid baselines such as BAGEL (7B MoT), 
\modelnamenc achieves a +6.9\% improvement on \textsc{MME-P} and +7.0\% on \textsc{MME-S}, indicating stronger perceptual and reasoning consistency. 
Against EMMA (4B), \modelnamenc further improves MMBench by +6.3\% and MathVista by +13.3\%, demonstrating superior mathematical and multi-step reasoning despite comparable model scale. 
Moreover, compared to the unified diffusion baseline MUDDIT, \modelnamenc achieves an overall improvement of 12\% across different understanding tasks. 
Finally, when compared with leading understanding-only models such as Qwen2.5-VL (3B), \modelnamenc attains 20.4\%  higher overall performance, underscoring the effectiveness of its pretrained VLM prior, bidirectional attention, and discrete diffusion formulation for multimodal understanding. 
Additional results on \textsc{OCRBenchV2}~\cite{fu2024ocrbench} can be found in Appendix~\ref{sec:supp_scene_text}.
Fig.~\ref{fig:img_und1} shows reasoning-based text generation examples, where \modelnamenc{} can accurately extract information from images to respond to user queries.
\Figunderstanding
\Figgenerationsimple

\subsection{Text-to-Image Generation}
Table~\ref{tab:combined_gen} summarizes the performance of \modelnamenc on
\textsc{GenEval} and \textsc{DPGBench} for multimodal generation. On \textsc{GenEval},
which evaluates compositional text-to-image generation across object
counting, attribute binding, and spatial reasoning, \modelnamenc achieves
an overall score of 0.95, outperforming strong unified baselines such as
EMMA and MammothModa2 by +2.2\% and +9.2\%,
respectively, highlighting
its stronger ability to associate attributes with the correct objects
under compositional constraints. 
A similar trend is observed on \textsc{DPGBench}, which evaluates fine-grained
prompt grounding across global understanding, attribute binding, and
relational reasoning, where \modelnamenc outperforms EMMA and
MammothModa2 by +6.5\% and +4.6\%, respectively. Notably, \modelnamenc
also surpasses generation-focused models such as Qwen-Image (7B+20B) by
4.0\% on \textsc{GenEval} and 3.2\% on \textsc{DPGBench}, despite using substantially fewer
parameters.  Fig.~\ref{fig:generation_1} further demonstrates that \modelnamenc
produces visually faithful and prompt-consistent images, accurately
rendering fine-grained details and background structures, which reflect
strong global semantics and local visual fidelity. 
\TabCombinedGenBench
\TabImgEditCombined

\subsection{Text-to-Image Editing}
Table~\ref{tab:image_edit_combined} summarizes the
image editing performance of \modelnamenc on \textsc{ImgEdit Bench}~\cite{ye2025imgedit}, \textsc{Emu-Edit}~\cite{sheynin2024emu}, and
\textsc{GEdit-Bench-EN}~\cite{liu2025step1x}. On \textsc{Emu-Edit}, \modelnamenc
outperforms EMMA and MammothModa2 by approximately +3.5\% and +4.1\%,
respectively, indicating stronger semantic alignment between the input
image, editing instruction, and edited output. On \textsc{GEdit-Bench-EN}, which
emphasizes perceptual quality and instruction satisfaction,
\modelnamenc improves the averaged score by +3.7\% over EMMA and +2.9\%
over MammothModa2.  Further, on \textsc{ImageEdit Bench}, which evaluates diverse editing scenarios including
object manipulation, background changes, style transfer, and hybrid
edits, \modelnamenc achieves an overall score of 4.24, surpassing EMMA
(4.01) and MammothModa2 (4.06) by +5.7\% and +4.4\%, respectively.
Notably, the largest gains are observed in Extract and Remove operations,
demonstrating more precise target isolation and reduced collateral
degradation. These improvements are driven by reference-based preference
alignment, which encourages \modelnamenc to select higher-quality edits
that better satisfy user intent. 
\Figeditingintelli


\noindent \textbf{Editing with reasoning.}
Fig.~\ref{fig:generation_1} presents image editing qualitative examples, where
\modelnamenc produces both accurate, large-scale semantic
edits (\eg style transfer) and fine-grained object-level modifications,
exhibiting strong instruction fidelity and precise edit localization.

Fig.~\ref{fig:edit_intelligent} further compares models on reasoning-driven editing tasks requiring temporal, geometric, and physical reasoning.
\modelnamenc generates outputs that better reflect the intended
transformations while preserving object identity, benefiting from the
strong reasoning priors inherited from the pretrained VLM backbone.

\Figgenemultiobject

\noindent \textbf{Subject-driven generation.}
Furthermore, \modelnamenc supports in-context subject-driven image generation from
multiple reference images, as shown in Fig.~\ref{fig:multi_object_gen},
without any explicit task-specific training. Given reference
images and a textual instruction, \modelnamenc synthesizes a coherent
output while preserving fine-grained visual details from the references.  
This behavior emerges from its unified multimodal optimization, which enables joint reasoning over
object identity, attributes, and spatial relations. 

\tableeditcomplexity

\noindent \textbf{Analysis on editing complexity.}  
To understand how reflection affects edits of varying difficulty, we conduct a complexity-controlled analysis on 200 images from PIE-Bench~\cite{ju2024pnp}, covering add/remove, reasoning-based, scene-text, and multi-object edits. Since reflection is generated in parallel with the edited output, this analysis isolates whether it provides useful localization and preservation cues across different edit types. As shown in Table~\ref{tab:reflection_editing}, reflection consistently improves PSNR and SSIM, reduces structural distance, and increases CLIP similarity for both edited and preserved regions.
The strongest gains appear on reasoning-based and multi-object edits, indicating that reflection is especially useful when the model must localize the correct region under more complex instructions.

\subsection{Ablations}
Table~\ref{tab:albtaions} presents a comprehensive ablation study analyzing the key design choices of \modelnamenc.

\noindent {\textbf{Model sizes.}}
Performance improves consistently as model size increases across all benchmarks. Larger backbones provide stronger multimodal priors and improved capacity for modeling long-range dependencies, which benefits both reasoning and diffusion-based generation. Notably, even the 4B model achieves competitive performance, validating the parameter-efficient design of \modelnamenc.

\noindent {\textbf{Visual tokenizer.}}
\modelnamenc uses PyraTok~\cite{susladkar2026pyratok}, which performs text-guided multi-scale quantization, enabling coarse-to-fine visual representations aligned with language. In contrast,  
3D-MBQ-VAE~\cite{susladkar2024motionaura} and MAGVIT-v2~\cite{yu2023language} use single-scale, visually trained tokenizers, limiting hierarchical modeling and text alignment. SweetTok~\cite{tan2025sweettok} incorporates text semantics but lacks multi-scale quantization, reducing its ability to capture coarse-to-fine structure.

\noindent {\textbf{Components.}}
Removing either understanding-specific or generation-specific LoRA adapters leads to noticeable degradation, confirming that separating task-specific adaptations is critical to avoid objective interference. Performance drops further when the router is removed, indicating that dynamic composition of adapters is necessary for balancing understanding and generation. Using a single shared LoRA fails entirely, demonstrating that naive parameter sharing causes severe entanglement between tasks.

\noindent {\textbf{Losses.}}
Removing visual or text alignment losses degrades performance on corresponding benchmarks.
Excluding reflection-based preference learning reduces editing and faithfulness metrics, showing that reasoning behind generation helps in precise instruction following and multimodal editing (refer to Appendix~\ref{sec:apn_mr} for visual results).

\TableAblationDVLM

\noindent \textbf{Alignment Training.} To assess the effect of our proposed mRef-DPO alingment method, we replace it with DPO~\cite{rafailov2023direct} or uni-GRPO~\cite{yang2025mmada}. Vanilla DPO can hurt when text–image tokens are weakly aligned, yielding noisy preference signals that degrade reasoning-grounded generation and edits. Uni-GRPO gives small gains but its group normalization is unstable (especially on short prompts), reducing fine-grained edit reliability. mRef-DPO performs best by using modality-aware preference learning to stabilize cross-modal credit assignment between textual reasoning and diffusion steps, improving alignment and edit precision across metrics.

\noindent \textbf{Time-step Conditioning.} In-context token concatenation is simple but lacks per-layer granularity, and its conditioning signal is diluted as sequence length grows (up to 4096 tokens in Stage~III), degrading DPGBench ($-1.10$) and ImgEdit ($-0.09$). AdLN underperforms as it overrides the frozen Qwen3-VL backbone's learned normalization statistics, disrupting its reasoning priors. In contrast, TSGN-RMSNorm modulates the existing RMSNorm affine parameters via zero-initialized multiplicative $(1 + s_\ell(t))$ and additive $b_\ell(t)$ terms, recovering the pretrained mapping exactly at $t{=}0$. This preserves the pretrained hidden-state geometry while injecting per-layer, timestep-dependent conditioning, with negligible overhead beyond the LoRA adapters.

\section{Conclusion}
We introduce UniDFlow, a unified vision--language diffusion model that performs understanding, text-to-image generation, and instruction-guided editing within a single discrete flow-matching framework. We further propose {mRef-DPO}, a reference-guided multimodal preference objective that jointly aligns text and image outputs relative to a frozen reference policy, improving faithfulness and controllability. Extensive results across six benchmarks show consistent gains, underscoring modality-aware preference alignment as critical for robust reasoning-grounded generation and precise visual editing.

\section*{Impact Statement}
This work presents a unified multimodal generative system that combines high-level understanding with high-fidelity visual generation. Such systems can enhance accessibility, creativity, and productivity by enabling natural multimodal interaction, supporting educational and design workflows, and improving human–computer interfaces. Our parameter-efficient training approach can also reduce computational cost compared to large-scale end-to-end retraining, potentially lowering environmental impact.

At the same time, improved generation and editing capabilities introduce risks. High-quality multimodal synthesis can be misused for deceptive media manipulation and precise editing may enable subtle alterations that are difficult to detect. Biases in pretrained vision–language backbones may also propagate into generated outputs, leading to stereotypical or harmful representations.

Our reference-based multimodal preference alignment in stage III aims to improve faithfulness and controllability by learning relative preferences under shared conditioning. This can help reduce spurious correlations and limit the amplification of dataset-specific artifacts when preference data is more balanced. However, alignment quality depends on the diversity and representativeness of the supervision signals, and misuse risks remain. Responsible deployment should therefore include safeguards such as content moderation, bias evaluation, and transparency mechanisms (\eg watermarking or provenance tracking).

\section*{Acknowledgments}
This research was partially supported by Google, the Google TPU Research Cloud (TRC) program, the National Science Foundation CAREER Award \#2542328, the U.S. Defense Advanced Research Projects Agency (DARPA) under award HR001125C0303, and the U.S. Army under contract W5170125CA160. The views and conclusions contained herein are those of the authors and should not be interpreted as necessarily representing the official policies, either expressed or implied, of Google, NSF, DARPA, the U.S. Army, or the U.S. Government. The U.S. Government is authorized to reproduce and distribute reprints for governmental purposes notwithstanding any copyright annotation therein.

\nocite{langley00}

\bibliography{main2}
\bibliographystyle{icml2026}

\newpage
\appendix

\onecolumn

\begin{figure*}[t!]
    \centering    \includegraphics[width=0.99\linewidth]{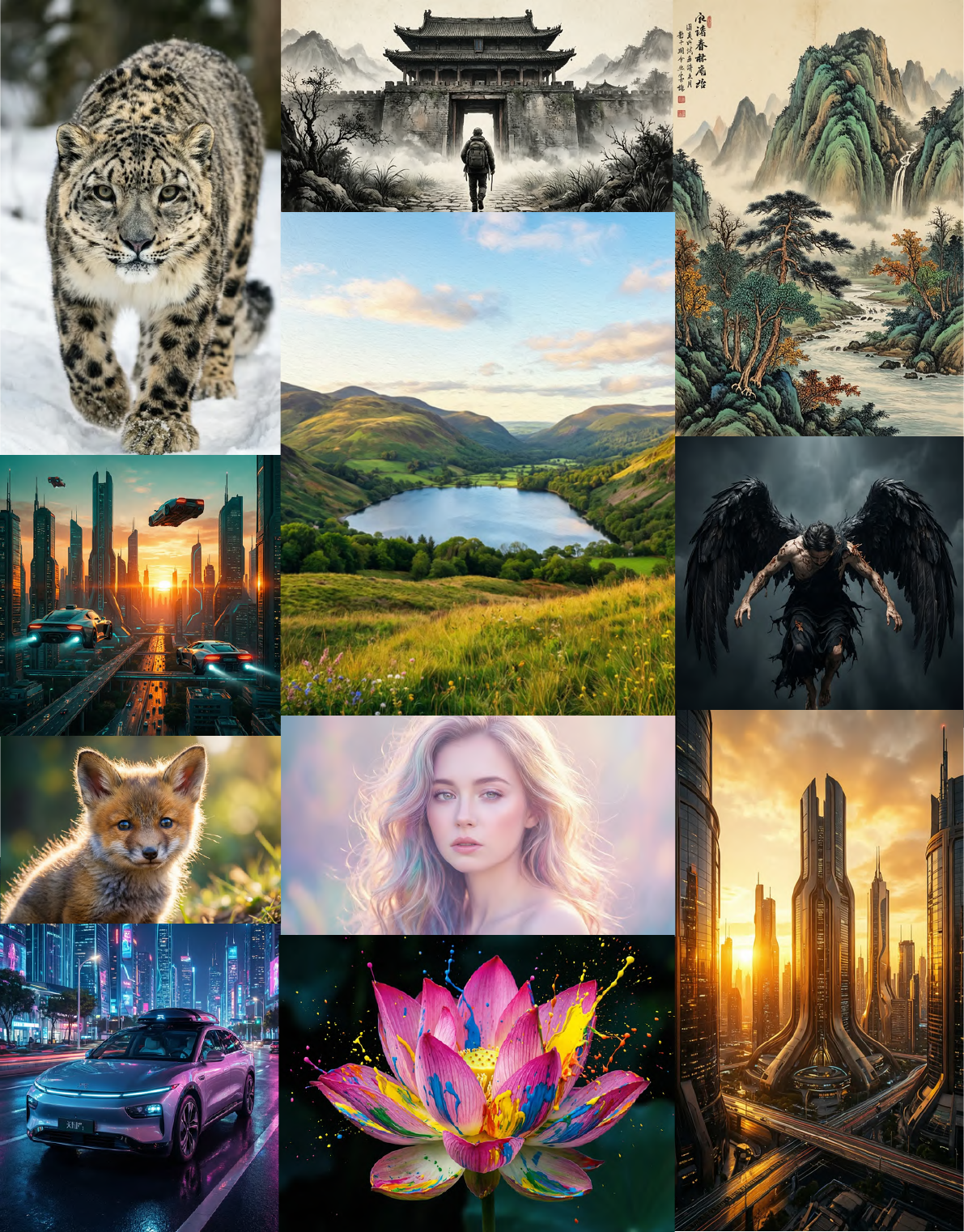}
    \caption{Image Generation with \modelnamenc}
    \label{fig:main_appendix}
\end{figure*}

\section{Implementation Details}
\label{sec:implementation}

We employ a three-stage training pipeline (Table~\ref{tab:implementation_details}) that progressively builds (i) visual instruction-following capability, (ii) high-fidelity visual generation, and (iii) joint multimodal understanding and alignment. Across all stages, we use AdamW optimization with mixed-precision training and gradient clipping to stabilize training at scale.

\textbf{Stage I: Text Alignment.}
We first perform supervised fine-tuning to teach the model to follow visual instructions and ground text responses in images. To improve robustness to real-world inputs, we train with variable aspect ratios and variable image resolutions, enabling the model to generalize across diverse image formats. The learning-rate schedule uses a warmup phase followed by cosine annealing for stable convergence. Additionally, we set $\lambda_{\mathrm{KL}} = 1.8$.

\textbf{Stage II: Visual Alignment.}
Next, we train the model for visual generation using a diffusion-based objective. We train at multiple resolutions (with variable aspect ratios) to encourage both global structure and fine detail, and use a linear learning-rate schedule with a longer warmup to support stable optimization under the generative objective. Regularization is applied via weight decay to improve generalization in this stage.

\textbf{Stage III: Reference-Based Multimodal Preference Alignment.}
Finally, we jointly optimize understanding and alignment, combining multimodal comprehension with aligned outputs. We expand the image-resolution range further and increase the maximum sequence length to support longer-context reasoning over visual content. This stage uses a cosine-annealed schedule with warmup and moderate regularization, aiming to consolidate gains from the first two stages while maintaining training stability at scale.

\begin{wrapfigure}{r}{0.54\linewidth}
  \vspace{-6pt}
  \centering
  \includegraphics[width=\linewidth]{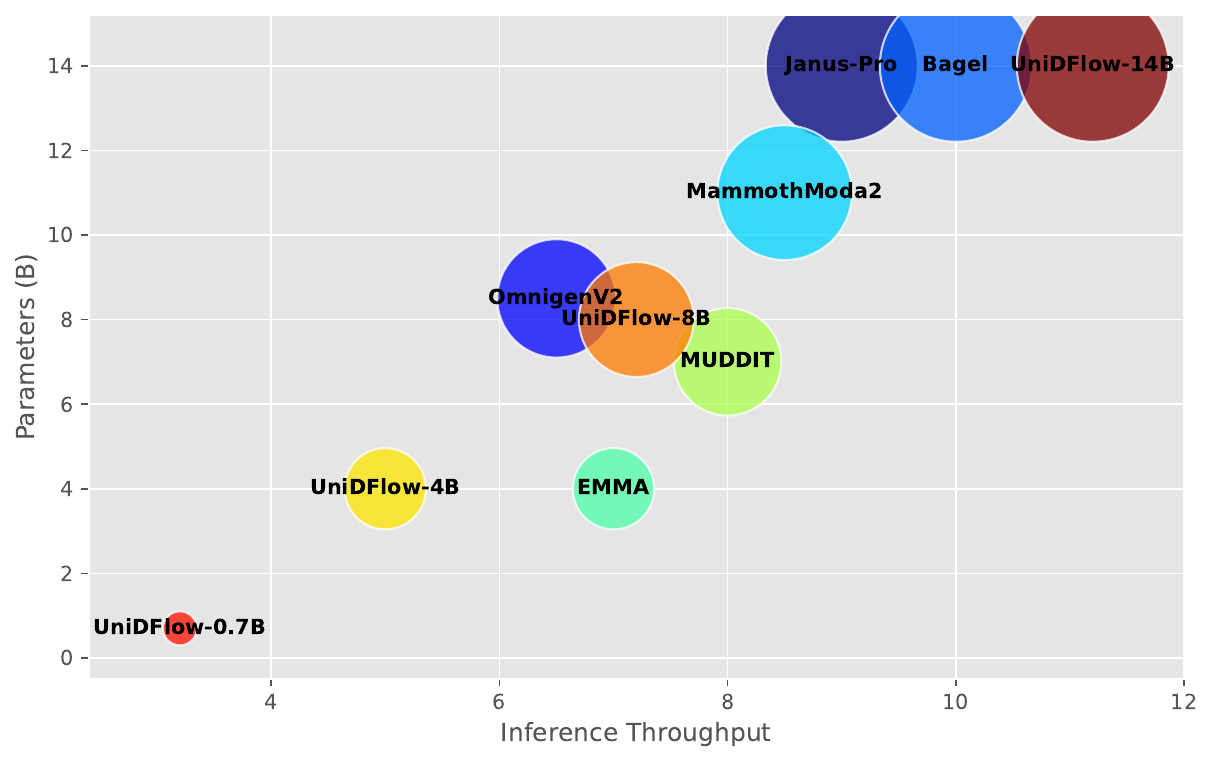}
  \vspace{-15pt}
  \caption{Inference throughput versus parameter count (in billions) for representative baselines and our model family. Higher throughput (right) is better, while fewer parameters (down) are more compact.}
  \label{fig:throughput_params}
  \vspace{-1pt}
\end{wrapfigure}

\textbf{Throughput--size trade-off.}
Figure~\ref{fig:throughput_params} summarizes the empirical efficiency landscape by plotting inference
throughput against model size for a set of representative systems (Janus-Pro~\cite{chen2025januspro}, OmniGenV2~\cite{wu2025omnigen2}, BAGEL~\cite{deng2025emerging},
MammothModa2~\cite{shen2025mammothmoda2}, EMMA~\cite{he2025emma}, and MUDDIT~\cite{shi2025muddit}) and our variants at 0.7B, 4B, 8B, and 14B parameters.
Rather than exhibiting a strictly monotonic dependence on parameter count, the scatter shows substantial
dispersion across independently implemented models, indicating that architectural choices and inference
stacks materially affect end-to-end throughput beyond raw scale.
In the large-model regime ($\sim$14B), \modelnamenc-14B attains the highest throughput among the compared
methods, outperforming other models of similar size (\eg BAGEL and Janus-Pro), suggesting improved
runtime efficiency at scale.
At intermediate sizes ($\sim$7--9B), \modelnamenc-8B is competitive with contemporaneous baselines, while the
smaller \modelnamenc-0.7B and \modelnamenc-4B provide lightweight operating points that prioritize compactness
with correspondingly lower throughput.

\section{Training Data}
\label{sec:data}

We employ a three-stage data curriculum that progressively transitions from supervised multimodal instruction learning to large-scale image-text pretraining and finally preference-based alignment for unified understanding, generation, and editing.  Table~\ref{tab:stage_data_summary} summarizes the training data.

\textbf{Token accounting.}
Throughout this work, the reported token counts include \emph{both} text tokens and discretized/embedded image tokens as consumed by the multimodal sequence interface (\ie the effective sequence length seen by the transformer). We report aggregate tokens per stage.

\textbf{Stage I: Text Alignment.}
In Stage~1, we initialize instruction-following behavior using MMInstruct~\cite{liu2024mminstruct}, a high-quality multimodal instruction tuning dataset spanning diverse domains and instruction types. We use approximately $1.0$M image--prompt--answer examples (MMInstruct reports $973$K instructions~\cite{liu2024mminstruct}) and train for $\approx 0.6$T total (image+text) tokens. 

\begin{wraptable}{r}{0.25\textwidth}
\centering
\caption{Stage-wise data summary (image+text tokens).}
\label{tab:stage_data_summary}
\resizebox{\linewidth}{!}{
\begin{tabular}{l l r r}
\toprule
\textbf{St.} & \textbf{Obj.} & \textbf{\#} & \textbf{Tok.} \\
\midrule
1 & SFT (MMInstruct) & 1.0M & 0.6T \\
2 & T2I (refined) & 4.5M & 1.2T \\
3 & Pref align. & 3.5M & 1.8T \\
\midrule
\multicolumn{3}{l}{\textbf{Total}} & 3.6T \\
\bottomrule
\end{tabular}
}
\end{wraptable}
\textbf{Stage II: Visual Alignment.}
Stage~2 focuses on \emph{text-to-image generative pretraining} to improve prompt adherence, compositional generalization, and broad visual coverage. We sample a total of $\approx 4.5$M images (with associated text prompts/captions) from three sources:
(i) $1.5$M from LAION-5B~\cite{schuhmann2022laion5b},
(ii) $1.0$M from JourneyDB~\cite{sun2023journeydb},
and (iii) $2.0$M from the \texttt{jackyhate/text-to-image-2M} collection on Hugging Face~\cite{jackyhate2024texttoimage2m}.
We refine and normalize the paired text using a proprietary LLM-based caption/prompt rewriting pipeline to reduce noise and increase instruction clarity, and train for $\approx 1.2$T total (image+text) tokens.

\textbf{Stage III: Reference-Based Multimodal Preference Alignment.}
Stage~3 aligns the model to high-quality, instruction-faithful outputs in our unified data format for (a) multimodal understanding, (b) image generation, and (c) image editing. We curate $\approx 3.5$M base tasks by aggregating:
OpenGPT-4o-Image ($\sim$80K)~\cite{chen2025opengpt4oimage},
AnyEdit-derived edits ($\sim$3.0M; AnyEdit reports 2.5M editing pairs)~\cite{yu2024anyedit},
and Pico-Banana-400K ($\sim$400K)~\cite{qian2025picobanana400k}.
We then convert these tasks into a high-quality preference dataset via rejection-sampling style annotation using proprietary multimodal LLMs:
for each edit instance, we generate and store (i) \emph{reflection} traces with a positive:negative ratio of $3:6$, and (ii) paired instruction/response candidates with positive:negative ratio $4:10$ (stored as accepted vs.\ rejected candidates in our format). Stage~3 consumes $\approx 1.8$T total tokens.

\begin{table}[t!]
\centering
\footnotesize
\caption{{Training setup by stage.} Stages I--III correspond to instruction tuning, visual generation, and joint understanding/alignment.}
\label{tab:implementation_details}
\setlength{\tabcolsep}{3.5pt}
\renewcommand{\arraystretch}{1.12}
\resizebox{0.75\linewidth}{!}{
\begin{tabular}{p{3.2cm}ccc}
\toprule
\textbf{Hyperparameter / Setting} &
\textbf{Stage I} &
\textbf{Stage II} &
\textbf{Stage III} \\
\midrule
GPUs & 32$\times$A100 (80GB) & 32$\times$H100 (80GB) & 48$\times$H100 (80GB) \\
Batch / GPU & 8 & 6 & 4 \\
Optimizer & AdamW & AdamW & AdamW \\
Init LR & $1\!\times\!10^{-5}$ & $5\!\times\!10^{-5}$ & $2\!\times\!10^{-5}$ \\
LR schedule & Cosine & Linear & Cosine \\
Warmup steps & 200 & 1000 & 1200 \\
Train steps & 10K & 25K & 30K \\
Grad accumulation & 6 & 4 & 4 \\
Max grad norm & 2.0 & 1.0 & 2.0 \\
Weight decay & 0 & $1\!\times\!10^{-2}$ & $1\!\times\!10^{-3}$ \\
Diffusion steps & 40 & 50 & 50 \\
Classifier-free guidance & 8 & 8 & 12 \\
Resolution & 224--1024  & 256/512/768/1024 & 224--1280  \\
Aspect ratios & 1:1, 4:3, 3:4 & 1:1, 16:9, 9:16 & 1:1, 4:3, 3:4, 16:9, 9:16 \\
Max seq length & 2048 & 2048 & 4096 \\
Precision & BF16 & BF16 & BF16 \\
GPU-Hours & 256 & 320 & 528 \\
\bottomrule
\end{tabular}
}
\end{table}

\begin{figure*}
  \centering
  \includegraphics[width=\textwidth]{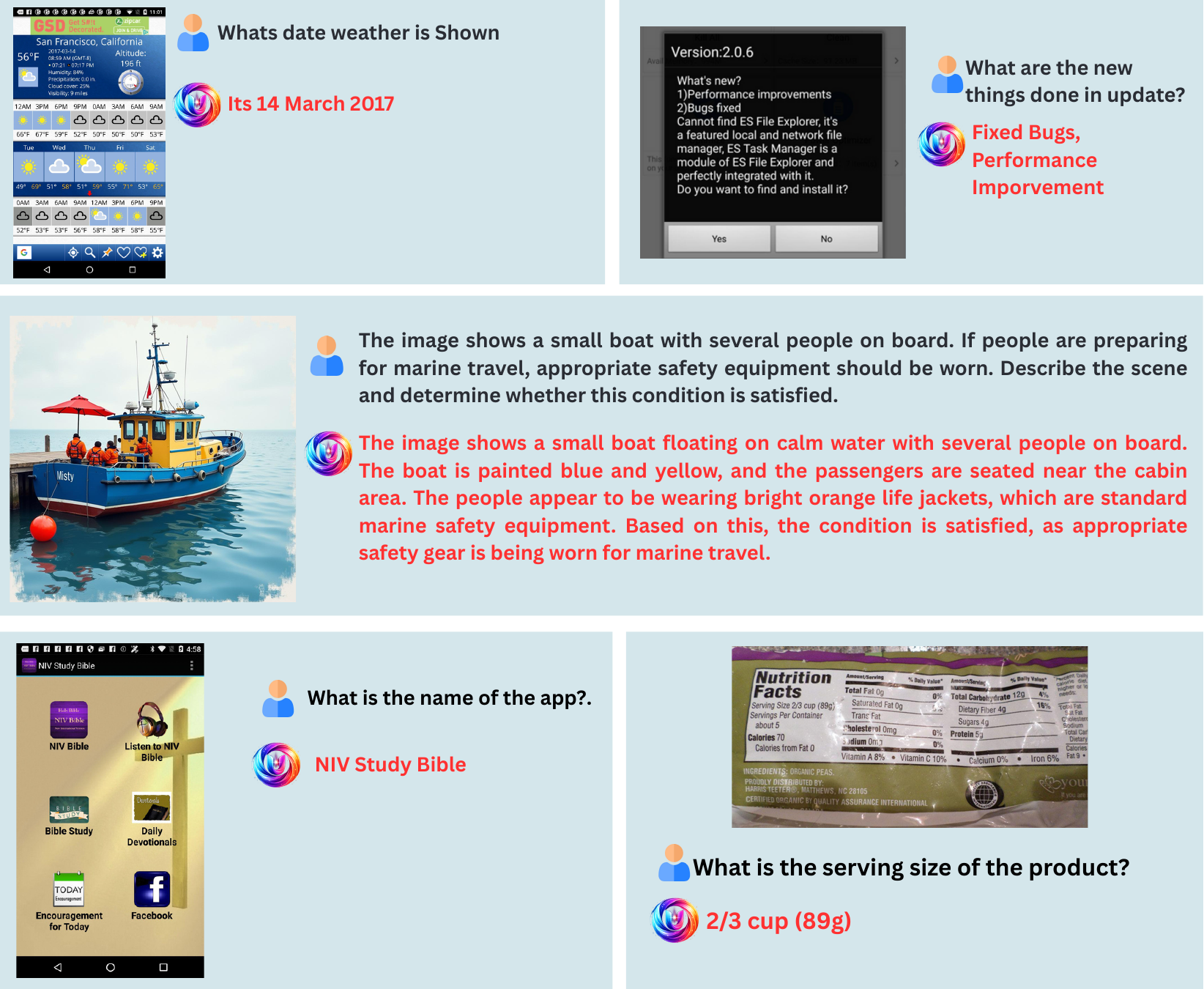}
  \caption{Image understanding and  reasoning with complex scenes.}
  \label{fig:und_supp1}
\end{figure*}

Table~\ref{tab:scene_text} reports accuracy on eight visual reasoning subtasks, Recognition, Referring, Spotting, Extraction, Parsing, Calculation, Understanding, and Reasoning, together with their macro Average on OCRBenchV2~\cite{fu2024ocrbench}. The compared systems include strong understanding-focused VLMs (\eg Qwen2.5-VL~\cite{bai2025qwen2}, InternVL~\cite{chen2024internvl}), unified understanding–generation models (\eg EMMA~\cite{he2025emma}, BAGEL~\cite{deng2025emerging}, MUDDIT~\cite{shi2025muddit}, MammothModa2~\cite{shen2025mammothmoda2}), and proprietary multimodal assistants (\eg GPT-4o, Gemini-Pro~\cite{google_gemini}). Fig.~\ref{fig:und_supp1} qualitatively shows \modelnamenc's robustness on complex understanding.  This evaluation is particularly diagnostic because it separates perceptual grounding (Recognition/Spotting/Extraction), structured interpretation (Parsing/Calculation), and holistic inference (Understanding/Reasoning).

\TableSceneText

Across model scales, our unified model family consistently dominates the subtask profile, with performance improving monotonically from Ours-4B → Ours-8B → Ours-14B. Concretely, Ours-14B achieves the best overall Average = 63.8, improving over the strongest baseline (MammothModa2, 56.1) by +7.7 points. Gains are broad rather than concentrated in a single capability: relative to the best previous work in every column, Ours-14B improves Recognition (76.7; +7.9), Referring (47.1; +7.6), Spotting (16.5; +2.8), Extraction (96.9; +4.3), Parsing (48.4; +9.2), Calculation (58.1; +7.9), Understanding (88.7; +8.5), and Reasoning (77.1; +9.0). Notably, the largest deltas occur in Parsing and Reasoning, suggesting that the proposed approach strengthens compositional/structured visual reasoning beyond raw perception.

A second takeaway is that even the compact variant (Ours-4B) is competitive with or better than substantially larger baselines: it reaches 58.4 Avg., exceeding MammothModa2 (56.1) and MUDDIT (54.7) while also improving the hardest “reasoning-heavy” columns (Calc. = 53.4, Reas. = 70.8). This aligns with the paper’s core design choice: rather than entangling understanding and generation in shared parameters, the method trains separate lightweight adapters for understanding vs. generation and combines them with a learned router, reducing objective interference and preserving specialization.

\section{Full Quantitative Results on GenEval and DPGBench}
\label{sec:full_quant_results}

The main paper reports the overall performance of \modelnamenc on GenEval and DPGBench. Here, we provide the complete attribute-wise breakdown used by both benchmarks. Table~\ref{tab:t2i_supp} shows that \modelnamenc achieves the best \emph{global} score and consistently improves across fine-grained categories (\eg entity, attribute, relation) as well as compositional criteria (single/two-object, counting, color, position, and color-attribute). These results indicate that the gains are not driven by a single subset of prompts; instead, \modelnamenc improves performance uniformly across evaluation dimensions, reflecting stronger text--image alignment and more reliable adherence to structured constraints.

\section{Additional Results}
\label{sec:apn_mr}
\begin{figure*}
  \centering
  \includegraphics[width=\textwidth]{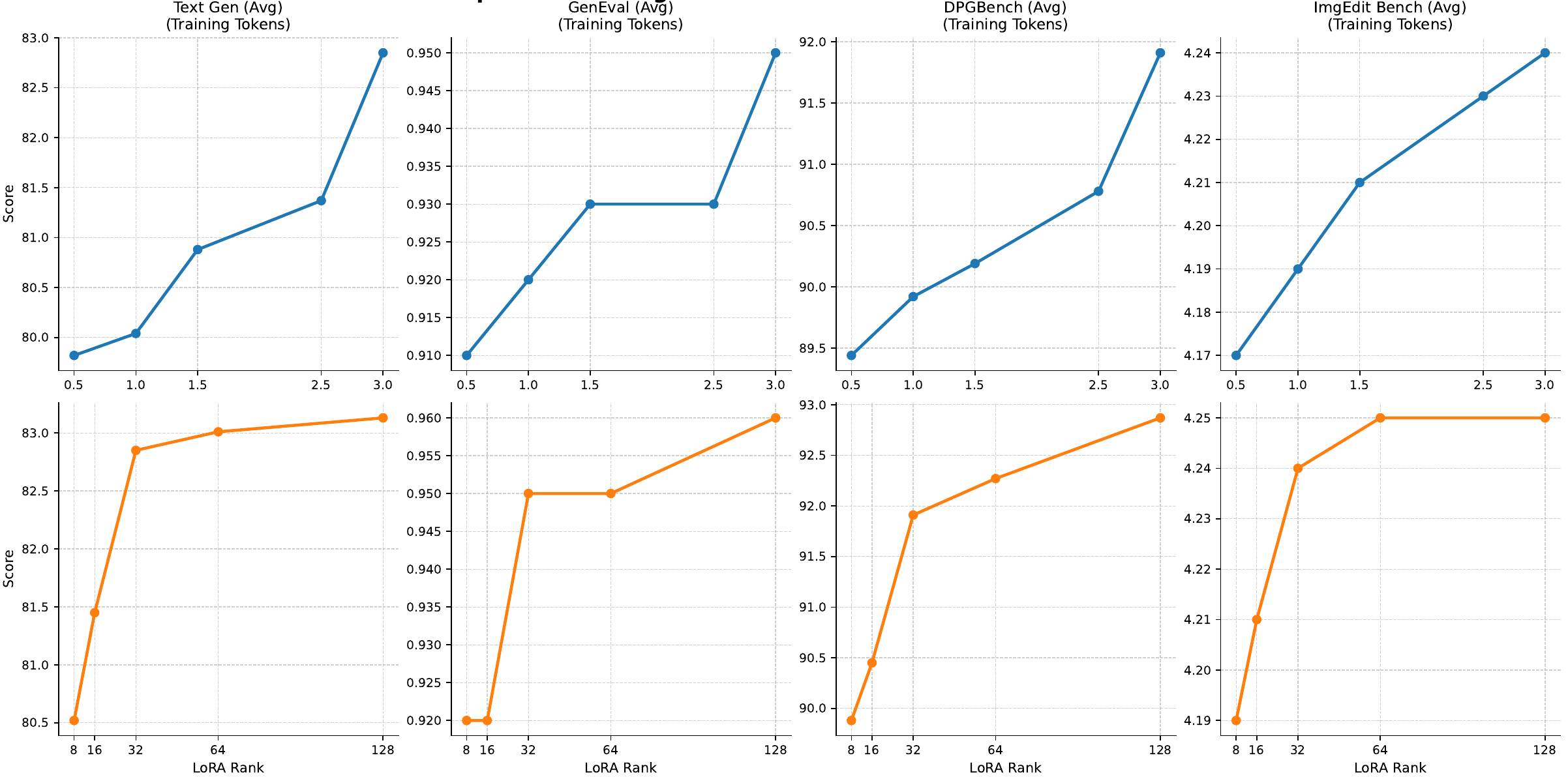}
  \caption{Analysis of the number of training tokens and LoRA ranks used across all stages.}
  \label{fig:abl_token_rank}
\end{figure*}

\noindent\textbf{Ablation on training tokens and LoRA rank.}
We study the effect of (i) the total number of pre-training tokens (image+text) and (ii) the LoRA rank used for adaptation. Figure~\ref{fig:abl_token_rank} shows a consistent improvement as we scale the training budget from 0.5T to 3T tokens, yielding substantial gains across \textit{TextGen}, \textit{GenEval}~\cite{ghosh2023geneval}, \textit{DPGBench}~\cite{hu2024ella}, and \textit{ImgEdit-Bench}~\cite{ye2025imgedit}. We also ablate the LoRA rank and observe that increasing the rank from 8 to 32 produces the largest marginal improvement across all benchmarks, indicating that low ranks under-parameterize the adaptation. Beyond rank 32, performance improvements diminish and largely saturate up to rank 128, suggesting the adaptation becomes capacity-sufficient. We use a default LoRA rank of 64 in all experiments.

\begin{figure*}
  \centering
  \includegraphics[width=\textwidth]{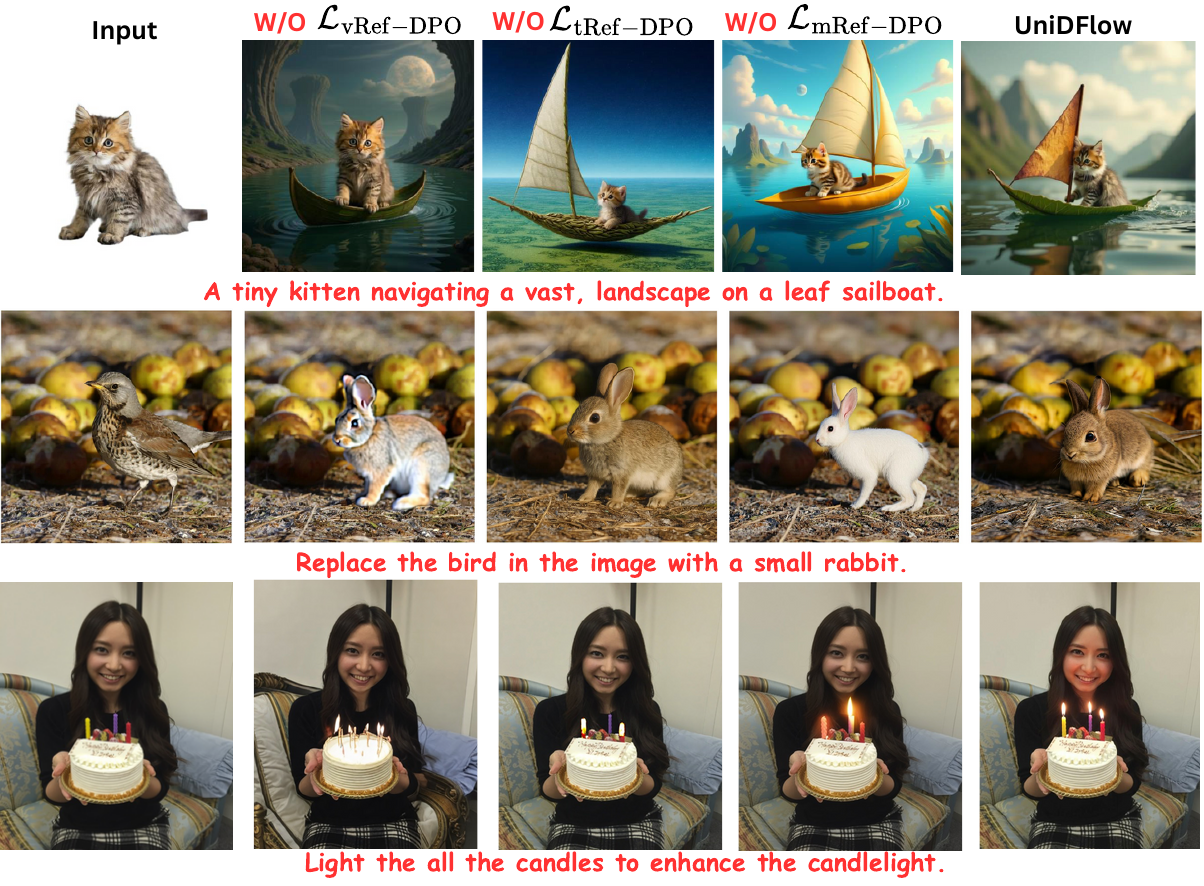}
  \caption{Visual comparison for Stage-III alignment losses.}
  \label{fig:abl_align_loss}
\end{figure*}

\noindent\textbf{Ablation on Stage-III losses.} Fig.~\ref{fig:abl_align_loss} compares the visual results for image editing without alignment losses.
Removing $\mathcal{L}_{\text{vRef-DPO}}$ weakens visual grounding. The model often preserves the coarse scene but fails to faithfully realize the requested edit, such as retaining the bird or producing incomplete candle lighting. Removing $\mathcal{L}_{\text{tRef-DPO}}$ degrades instruction following, leading to missing or semantically incorrect target objects, \eg the kitten is not properly placed on the leaf sailboat and the rabbit replacement becomes unreliable. Removing $\mathcal{L}_{\text{mRef-DPO}}$ introduces stronger cross-modal inconsistencies and visual artifacts, such as unrealistic object appearance, poor scene integration, or excessive color changes. In contrast, the full UniDFlow model better preserves the input context while executing the requested edit, producing a kitten plausibly navigating a leaf sailboat, replacing the bird with a scene-consistent rabbit, and lighting the candles without distorting the subject. These results show that the three alignment losses are complementary, with visual-reference alignment improving preservation and grounding, text-reference alignment improving semantic edit fidelity, and multimodal reference alignment stabilizing cross-modal credit assignment for more reliable reasoning-grounded generation.
\TableSuppGen

\section{Extended Related Work}\label{sec:extended_related_work}

\noindent \textbf{Diffusion for Visual Generation.}
Diffusion probabilistic models (DPMs)~\cite{ho2020denoising,nichol2021glide,saharia2022photorealistic} surpass GANs~\cite{goodfellow2014generative} in stability and generation quality, but are computationally expensive due to pixel-space diffusion. Latent diffusion models (LDMs)~\cite{rombach2022high} mitigate this cost by operating in a compressed latent space and achieve strong text-to-image performance~\cite{zhang2023adding,chen2023pixart,podell2023sdxl}. However, continuous Gaussian diffusion is well-suited for images but less natural for discrete modalities such as text. Discrete diffusion~\cite{austin2021structured} addresses this by using categorical corruption (\eg masking), motivating image generators that replace autoregressive decoding with parallel mask-and-predict refinement, improving both fidelity and latency~\cite{gu2022vector,chang2022maskgit}. 

\noindent \textbf{LLMs and VLMs for Understanding.}
Large language models (LLMs)~\cite{touvron2023llama,liu2024deepseek} have achieved strong zero-shot reasoning and instruction-following by autoregressively generating tokens with decoder-only Transformers~\cite{vaswani2017attention}. Inspired by their success, vision–language models (VLMs)~\cite{bai2023qwen,bai2025qwen2} extend LLMs to visual inputs by coupling a vision encoder (\eg SigLIP~\cite{zhai2023sigmoid}) with a language model via lightweight projection layers, treating images as sequences of visual tokens. Models such as Qwen~\cite{bai2025qwen3vl}, LLaVA~\cite{liu2023visual}, BLIP-2~\cite{li2023blip}, and Flamingo~\cite{alayrac2022flamingo} enable strong visual understanding (\eg captioning, VQA), but treat vision as read-only and rely on separate diffusion models for image generation.

Beyond likelihood training, preference alignment has been extended from LLMs to diffusion models using DPO-style objectives~\cite{rafailov2023direct}. Diffusion-DPO~\cite{wallace2024diffusion} directly fine-tunes text-to-image models on pairwise human judgments via a likelihood-based preference loss, while DSPO~\cite{zhu2025dspo} instead aligns the diffusion score function in score space, staying closer to the original training objective. Subsequent variants such as DGPO~\cite{luo2025reinforcing} improve stability through group-wise preference optimization, and recent work further generalizes DPO-style alignment to discrete diffusion processes~\cite{borso2025preference}, bridging continuous and categorical diffusion frameworks.

\noindent \textbf{Unified Models for Understanding and Generation.}
Diffusion models and vision–language models excel at generation and semantic understanding, respectively, motivating unified architectures. To improve generation and editing, recent models add additional alignment stages. OmniGenV2~\cite{wu2025omnigen2} employs multimodal reflection for self-correction, while MammothModa2~\cite{shen2025mammothmoda2} applies reinforcement learning with scalar rewards (\eg OCR and aesthetic scores). 
In contrast, \modelnamenc introduces reference-based preference alignment across text and vision with reflection, enabling stable and faithful generation and editing.

\begin{figure*}
  \centering
  \includegraphics[width=\textwidth]{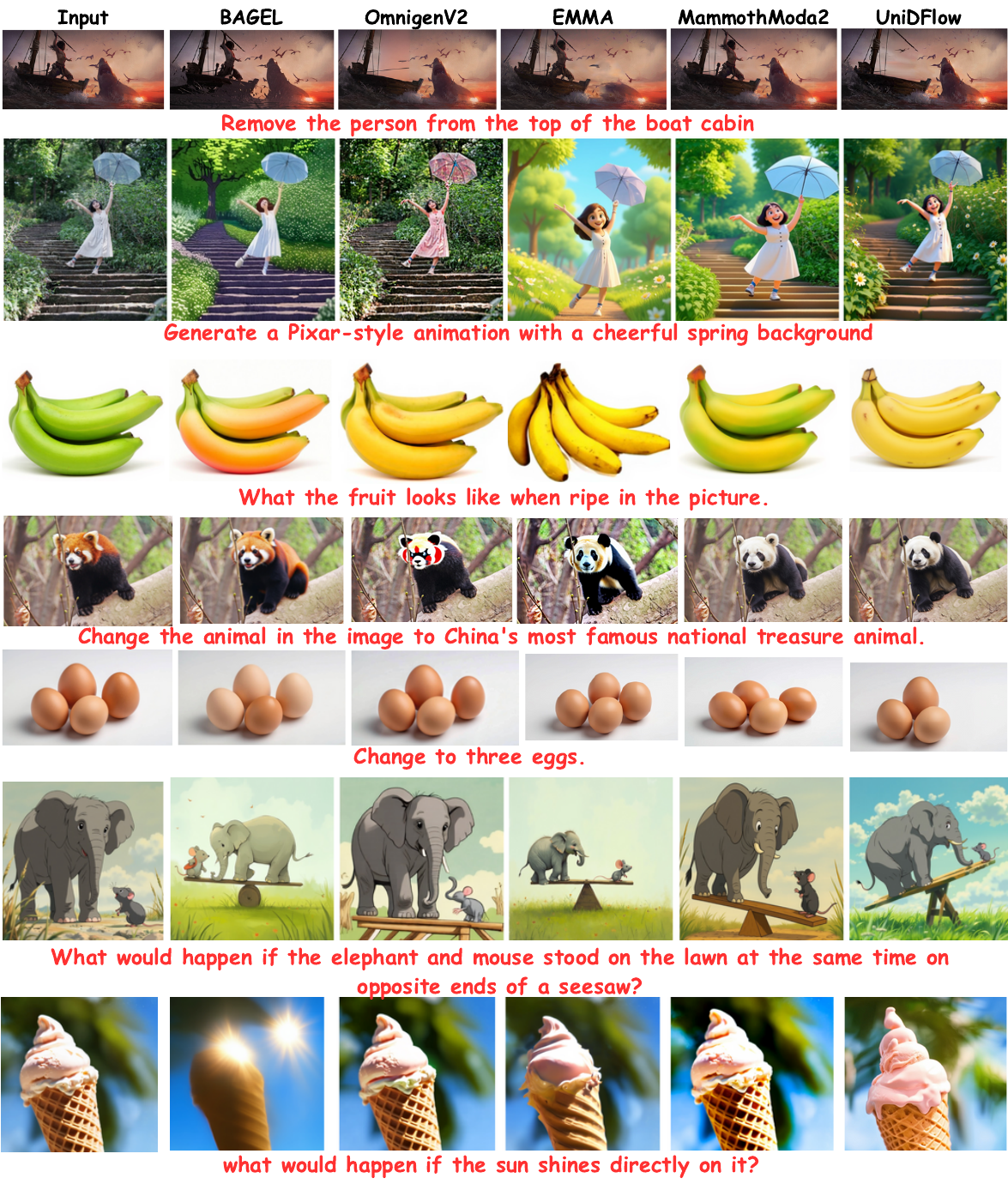}
  \caption{Qualitative comparison of reasoning-based image editing on complex images. Given an input image and a natural-language instruction, each column shows the result produced by a different editing method. The examples highlight challenging edits that require semantic understanding and commonsense reasoning, including object removal, style transformation, ripeness prediction, species replacement, quantity modification, physical interaction reasoning, and sunlight-induced melting effects.}
  \label{fig:compare_supp1}
\end{figure*}

\begin{figure*}
  \centering
  \includegraphics[width=\textwidth]{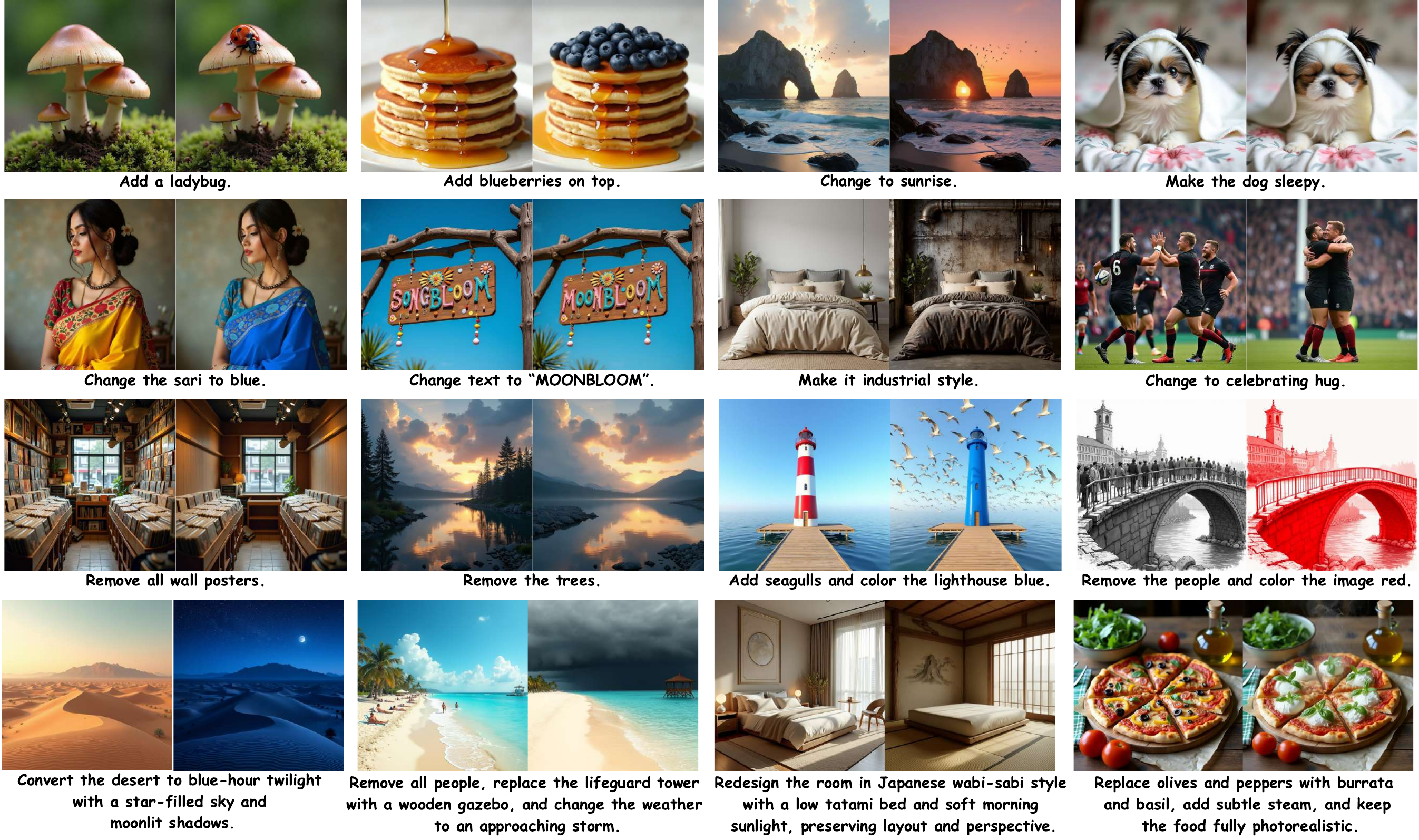}
  \caption{Image editing results on complex scenarios involving object, style, text, color, weather, and layout transformations.}
  \label{fig:supp_more_edit4}
\end{figure*}

\begin{figure*}[t]
  \centering
  \includegraphics[
    width=\textwidth,
    height=\textheight,
    keepaspectratio,
  ]{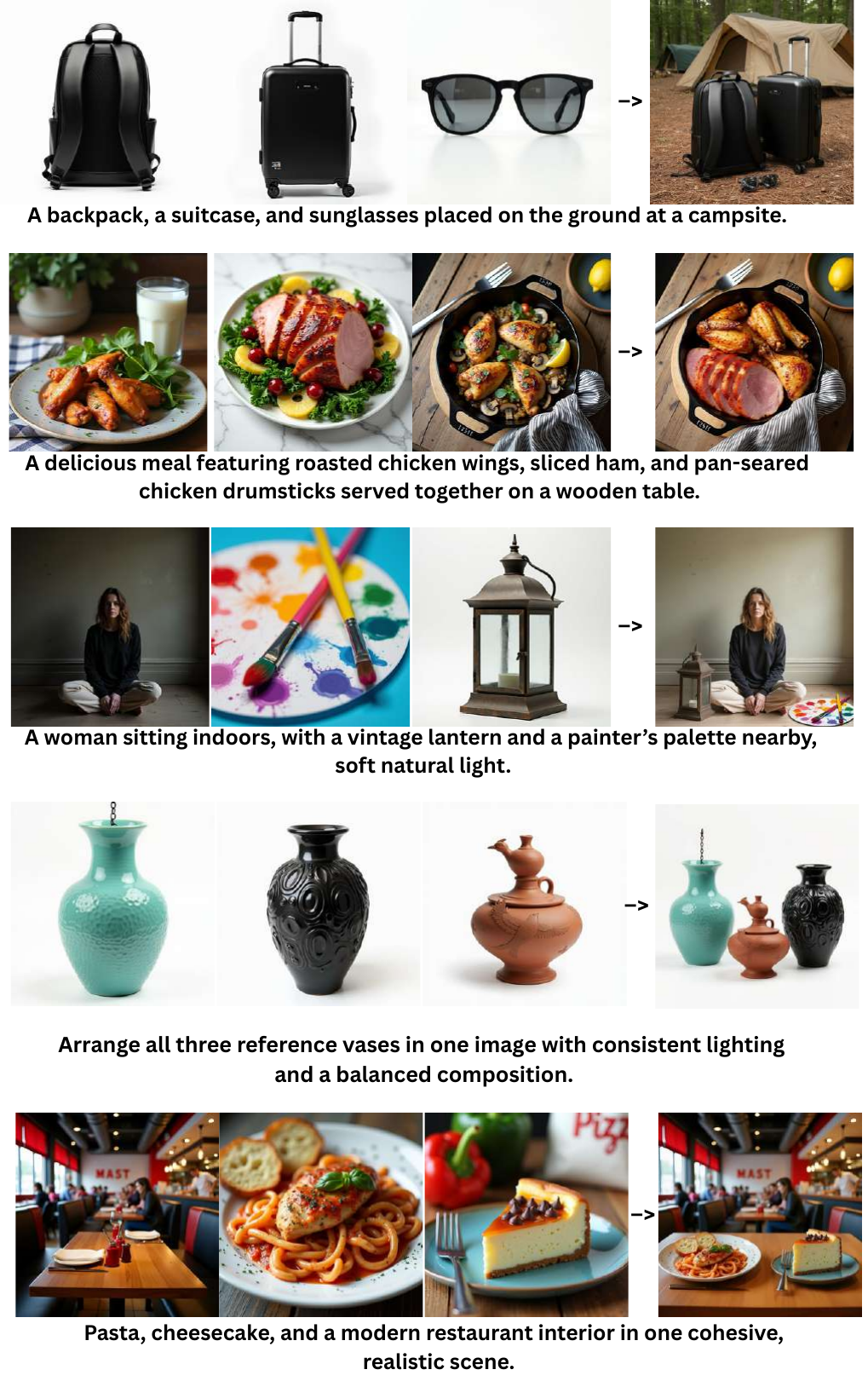}
  \caption{Zero-shot multi-subject reasoning-based editing.}
  \label{fig:fullsize}
\end{figure*}

\begin{figure*}[t]
  \centering
  \includegraphics[
    width=\textwidth,
    height=\textheight,
    keepaspectratio,
  ]{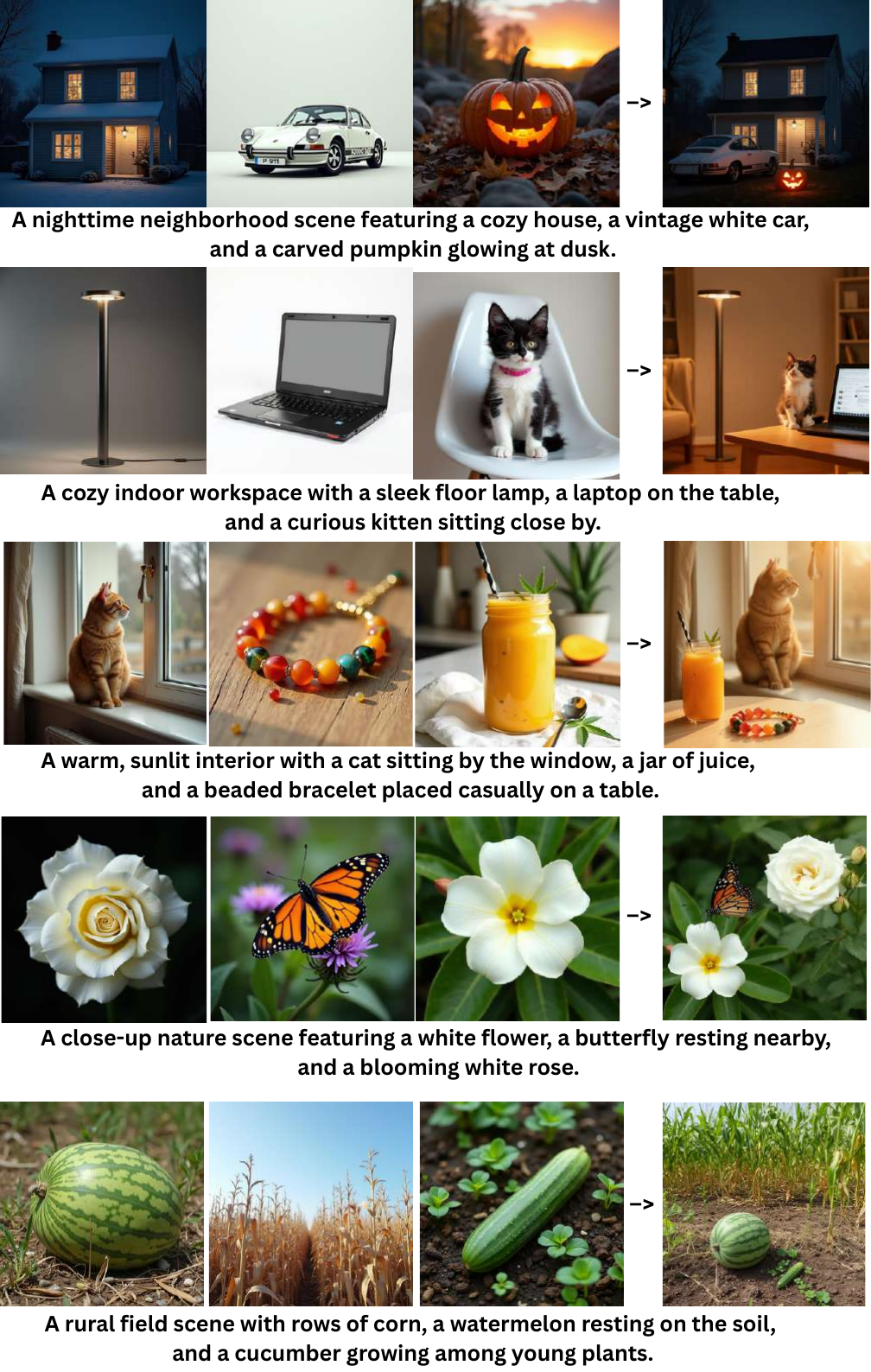}
  \caption{Zero-shot multi-subject reasoning-based editing.}
  \label{fig:fullsize2}
\end{figure*}

\begin{figure*}[t]
  \centering
  \includegraphics[
    width=\textwidth,
    height=\textheight,
    keepaspectratio,
  ]{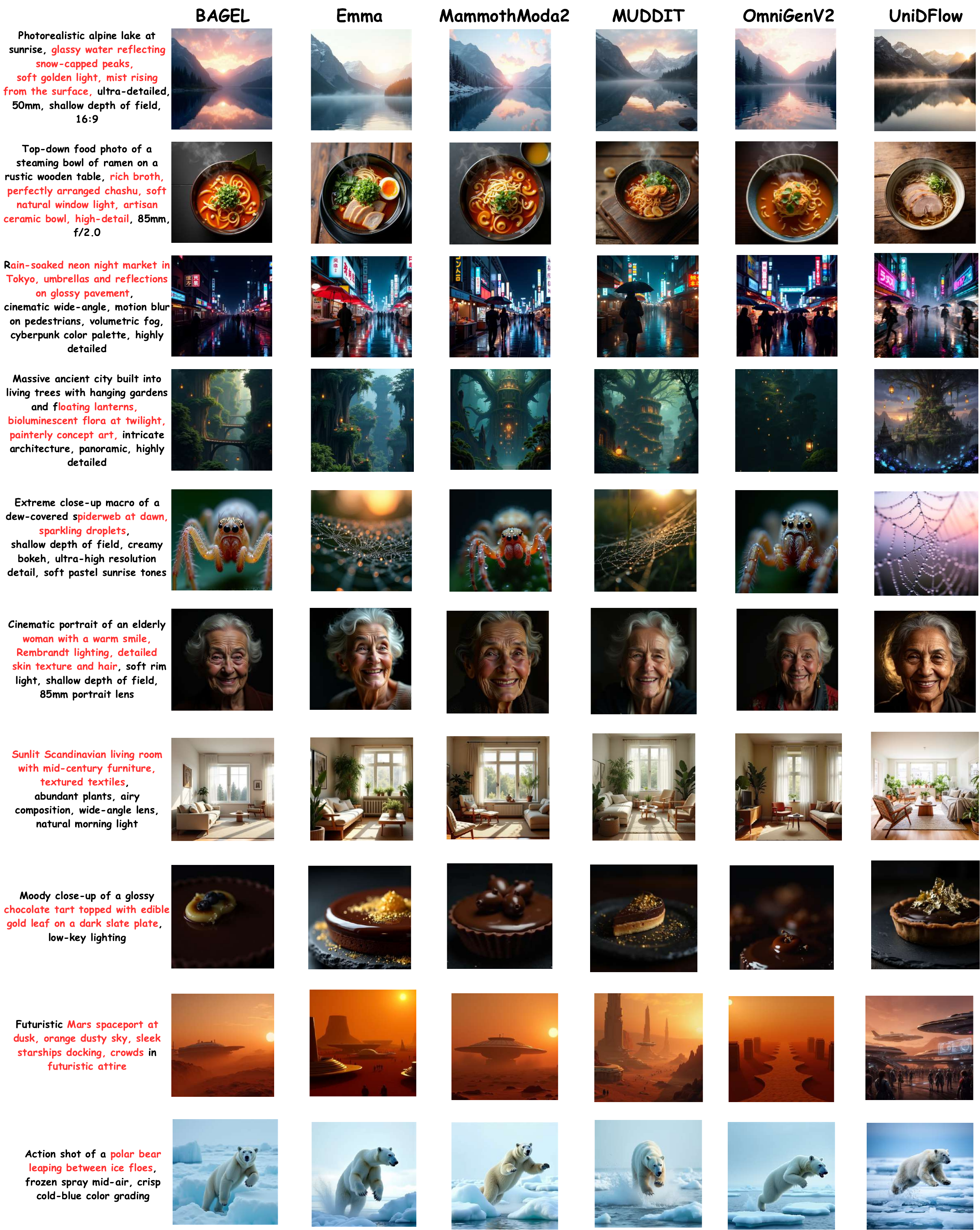}
  \caption{Text-to-image generation comparison against baselines.}
  \label{fig:T2ISupp}
\end{figure*}

\begin{figure*}
  \centering
  \includegraphics[width=\textwidth, height=\textheight]{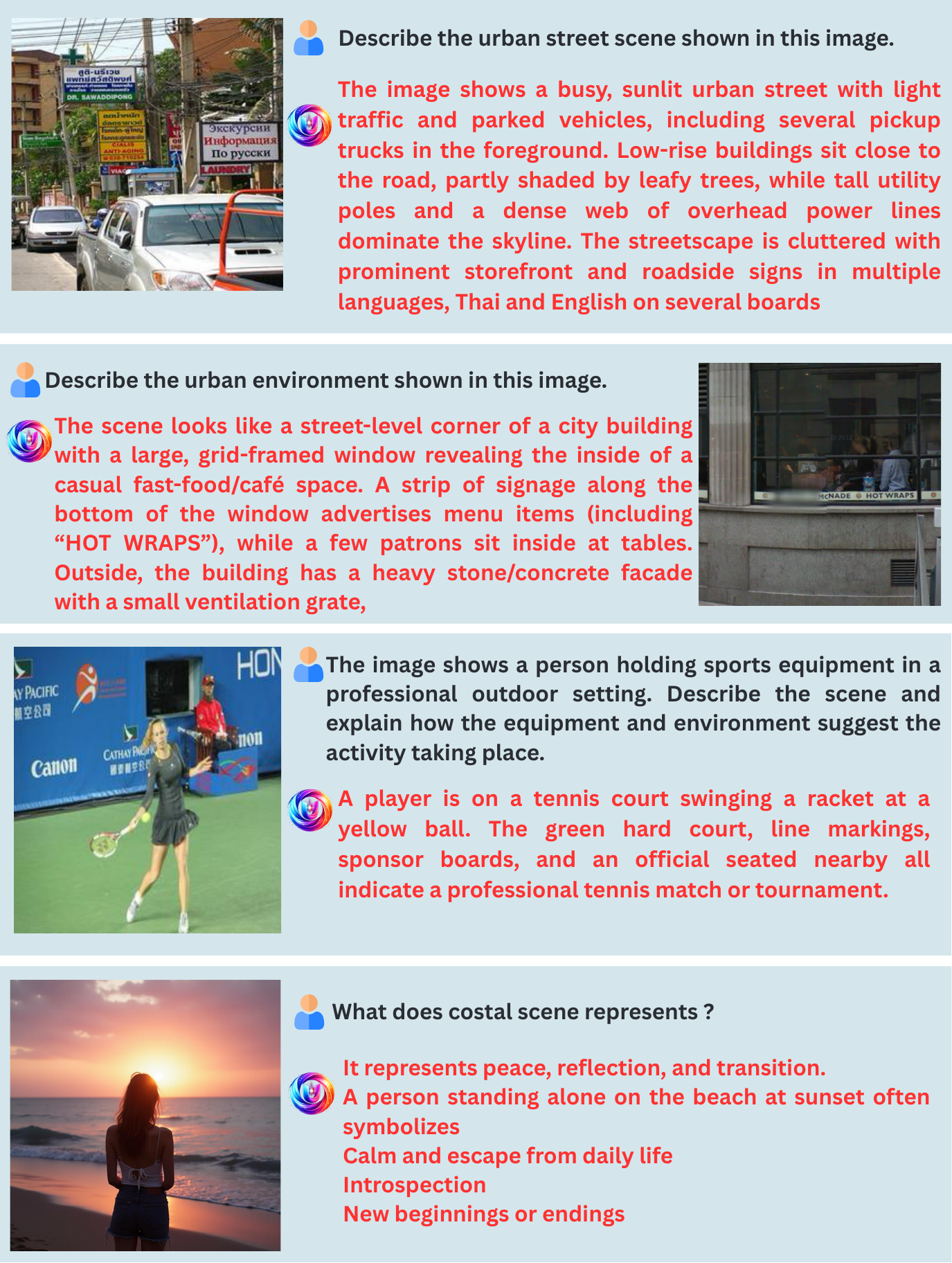}
  \caption{Image-to-text generated results with \modelnamenc.}
  \label{fig:und_supp2}
\end{figure*}

\begin{figure*}
  \centering
  \includegraphics[width=\textwidth]{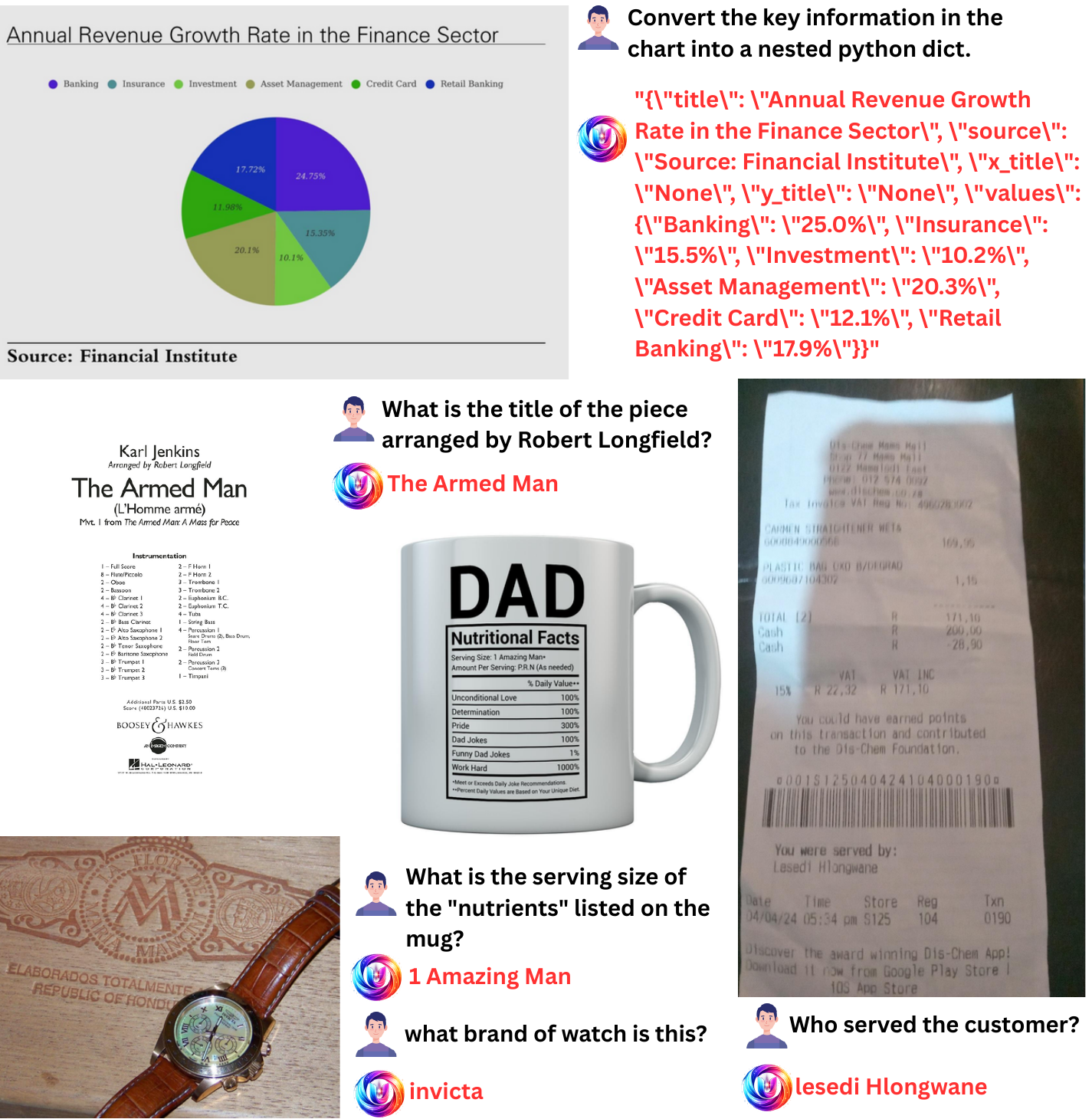}
  \caption{Image-to-text generated results with \modelnamenc.}
  \label{fig:und_supp3}
\end{figure*}

\begin{figure*}
  \centering
  \includegraphics[width=\textwidth]{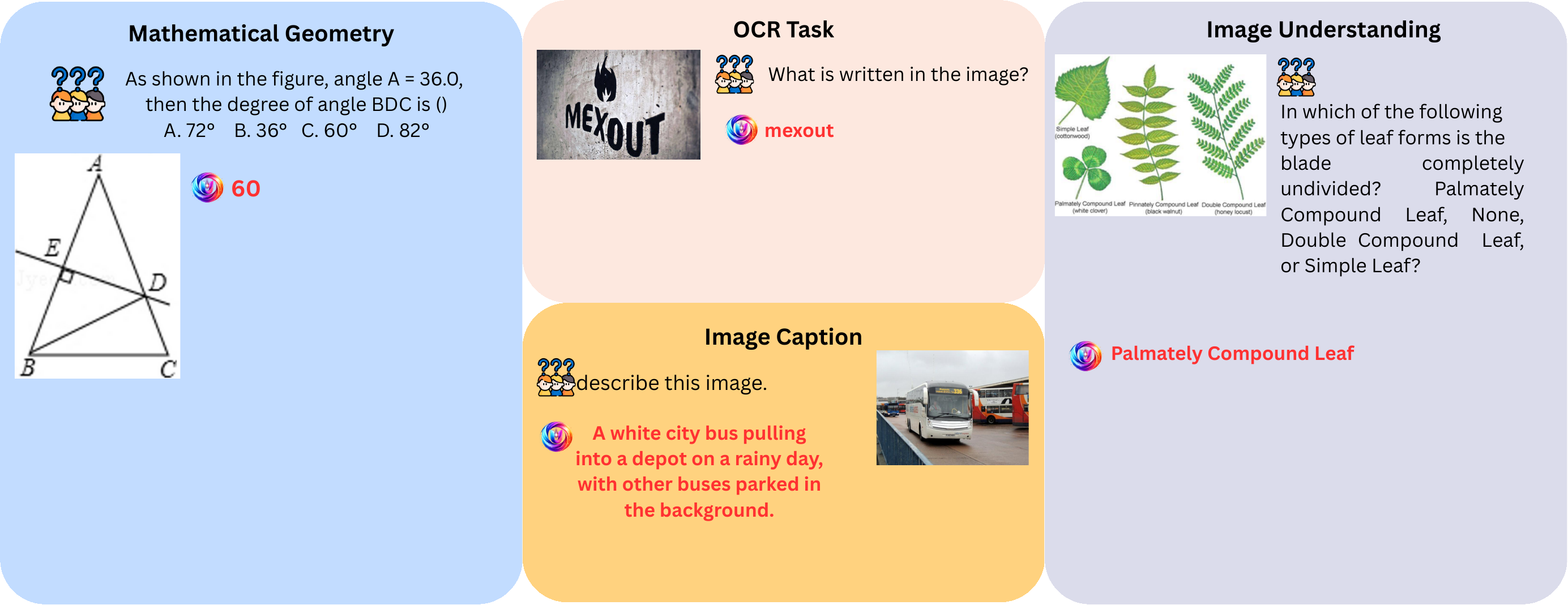}
  \caption{More complex reasoning tasks with \modelnamenc.}
  \label{fig:und_supp5}
\end{figure*}

\begin{figure*}
  \centering
  \includegraphics[width=\textwidth]{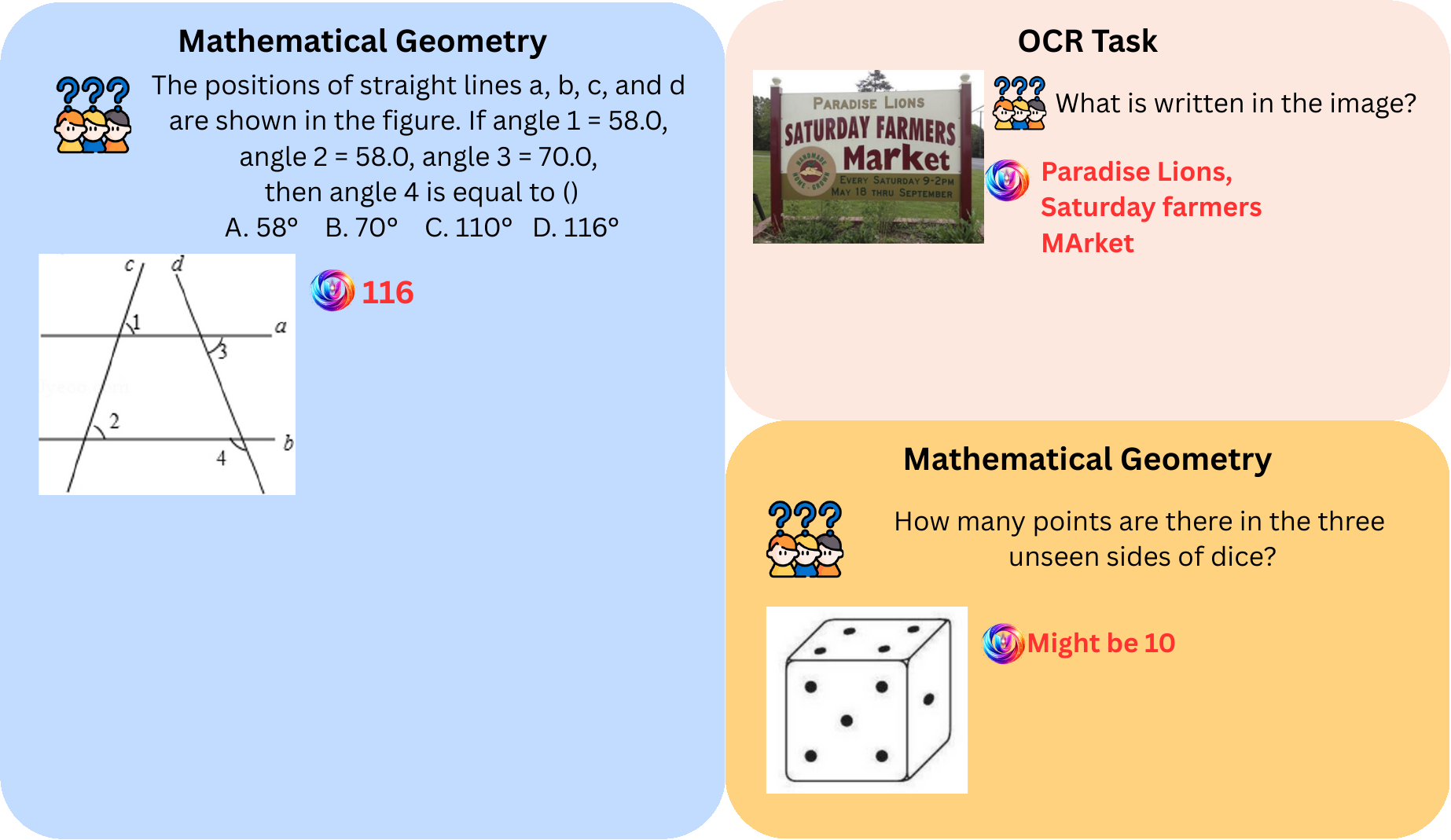}
  \caption{More complex reasoning tasks with \modelnamenc.}
  \label{fig:und_supp5}
\end{figure*}

\begin{tcolorbox}[
  title={Dataset curation prompt used in our pipeline},
  colback=white,
  colframe=black,
  boxrule=0.4pt,
  arc=2pt,
  left=2mm,right=2mm,top=1mm,bottom=1mm
]
\begin{Verbatim}[
  fontsize=\scriptsize,
  baselinestretch=0.92,
  breaklines=true,
  breakanywhere=true,
  breaksymbolleft={},
  breaksymbolright={}
]
You are an expert data curator for multimodal image-editing instruction datasets.

Inputs:
1) SOURCE_IMAGE: the original image
2) RAW_INSTRUCTION: the original instruction text
3) EDITED_IMAGE: the ground-truth edited image

Tasks:
1) Produce clean training fields.
2) Create preference pairs by writing one "chosen" output and >=3 "rejected" alternatives (plausible but worse).

Rules:
- Preserve intent; if ambiguous, pick the interpretation that matches EDITED_IMAGE.
- Be consistent with what is visible in SOURCE_IMAGE and EDITED_IMAGE; do not invent unseen details.
- Be explicit: what changes, where, how much/style constraints, and what must remain unchanged if relevant.
- Reflection describes the final edited image relative to the original; only observable outcomes.

Rejected alternatives (>=3):
- Each must be a realistic mistake (not nonsense) and include:
  - worse tuned instruction
  - worse edit directive
  - worse reflection of the incorrect outcome
  - "why_rejected" explaining what is wrong vs the chosen
  - "negative_type" from: under-edit, over-edit, wrong-attribute, wrong-region, unwanted-add/remove
- Include at least one near-miss (almost correct but subtly wrong: intensity, shade, lighting, missed constraint).

Output JSON (return ONLY valid JSON; exactly these keys):
{
  "prompt": string,
  "image": "SOURCE_IMAGE",
  "answer_instruction_tuned": string,
  "edit_instruction_for_image": string,
  "edited_image": "EDITED_IMAGE",
  "reflection_of_edited_image": string,
  "preference_data": {
    "prompt": string,
    "chosen": {
      "answer_instruction_tuned": string,
      "edit_instruction_for_image": string,
      "reflection_of_edited_image": string
    },
    "rejected": [
      {
        "answer_instruction_tuned": string,
        "edit_instruction_for_image": string,
        "reflection_of_edited_image": string,
        "why_rejected": string,
        "negative_type": string
      }
    ]
  }
}

Now process:
SOURCE_IMAGE: <<IMAGE>>
RAW_INSTRUCTION: <<INSTRUCTION_TEXT>>
EDITED_IMAGE: <<EDITED_IMAGE>>
\end{Verbatim}
\end{tcolorbox}

\label{sec:supp_scene_text}

\end{document}